\documentclass[fullpaper,final]{nldl}

\paperID{42}

\vol{233}

\usepackage{mathtools}
\usepackage{amsmath}
\usepackage{amssymb}

\usepackage{graphicx}

\usepackage{booktabs}

\usepackage{enumitem}

\usepackage{algorithm}

\usepackage{algpseudocode}

\usepackage{listings}
\usepackage{hyperref}
\usepackage{url}
\hypersetup{
  colorlinks,
  linkcolor = BrickRed,
  citecolor = NavyBlue,
  urlcolor  = Magenta!80!black,
}

\usepackage[toc,page,header]{appendix}
\usepackage{minitoc}

\title{Hubness Reduction Improves Sentence-BERT Semantic Spaces}
\author[1]{Beatrix M. G. Nielsen\thanks{Corresponding Author.}}
\author[1]{Lars Kai Hansen}
\affil[1]{Technical University of Denmark}
\affil[ ]{\texttt{\{bmgi, lkai\}@dtu.dk}}

\begin{document}
\doparttoc 
\faketableofcontents 

\maketitle

\begin{abstract}
Semantic representations of text, i.e. representations of natural language which capture meaning by geometry, are essential for areas such as information retrieval and document grouping. High-dimensional trained dense vectors have received much attention in recent years as such representations. We investigate the structure of semantic spaces that arise from embeddings made with Sentence-BERT and find that the representations suffer from a well-known problem in high dimensions called hubness. Hubness results in asymmetric neighborhood relations, such that some texts (the hubs) are neighbours of many other texts while most texts (so-called anti-hubs), are neighbours of few or no other texts. We quantify the semantic quality of the embeddings using hubness scores and error rate of a neighbourhood based classifier. We find that when hubness is high, we can reduce error rate and hubness using hubness reduction methods. We identify a combination of two methods as resulting in the best reduction. For example, on one of the tested pretrained models, this combined method can reduce hubness by about 75\% and error rate by about 9\%. Thus, we argue that mitigating hubness in the embedding space provides better semantic representations of text.
\end{abstract}

\section{Introduction}
Large Language Models (LLMs) like BERT \citep{devlin-etal-2019-bert} have been shown to be effective for information retrieval (e.g. \citep{karpukhin-etal-2020-dense}, \citep{jiang-etal-2020-cross}) and using the siamese network construction, the Sentence-BERT architecture \citep{reimers-gurevych-2019-sentence} has made it possible to create semantically meaningful text embeddings effectively enough to make large-scale text clustering viable. Sentence-BERT can be viewed as a fine tuning of a LLM to produce semantically relevant sentence embeddings. We explore the geometry of the resulting embedding space using the neighbourhoods of the data points. \par
For a typical Sentence-BERT setup, embeddings are of dimension $768$. In such high dimensional data, so-called hubs tend to occur (e.g., \citep{Beyer1999}, \citep{radovanovic2010hubs}). Hubs are points which are among the $k$ nearest neighbours of many other points even for small $k$. Antihubs, which are in the nearest neighbours of few or no other points, also tend to appear in large numbers when the dimension is high ($\gtrsim$ 10). These hubs and antihubs skew the nearest neighbour relation, since if a point, $x$, is in the ten nearest neighbours of a hundred other points, most of those points cannot be in the ten nearest neighbours of $x$. Therefore, if we use the neighbour relation to define semantic similarity, we get the counter intuitive result, that some texts are semantically similar to many other texts while many texts are not semantically similar to anything. Hubness has been found in many different kinds of data and degrades performance of neighbourhood based algorithms whether they are used for classification \citep{radovanovic2010hubs}, regression \citep{buza2015nearest}, clustering (e.g. \citep{schnitzer2015unbalancing}, \citep{tomasev2013role}) or outlier detection \citep{radovanovic2014reverse}. Hubness has also been found to cause vulnerability in recommender systems \citep{hara2015reducing}. Naturally, there has been a quest in the cited literature for means to mitigate the effect of hubness. \par 

\begin{figure*}[htb]
\centering
\includegraphics[width=0.9\linewidth]{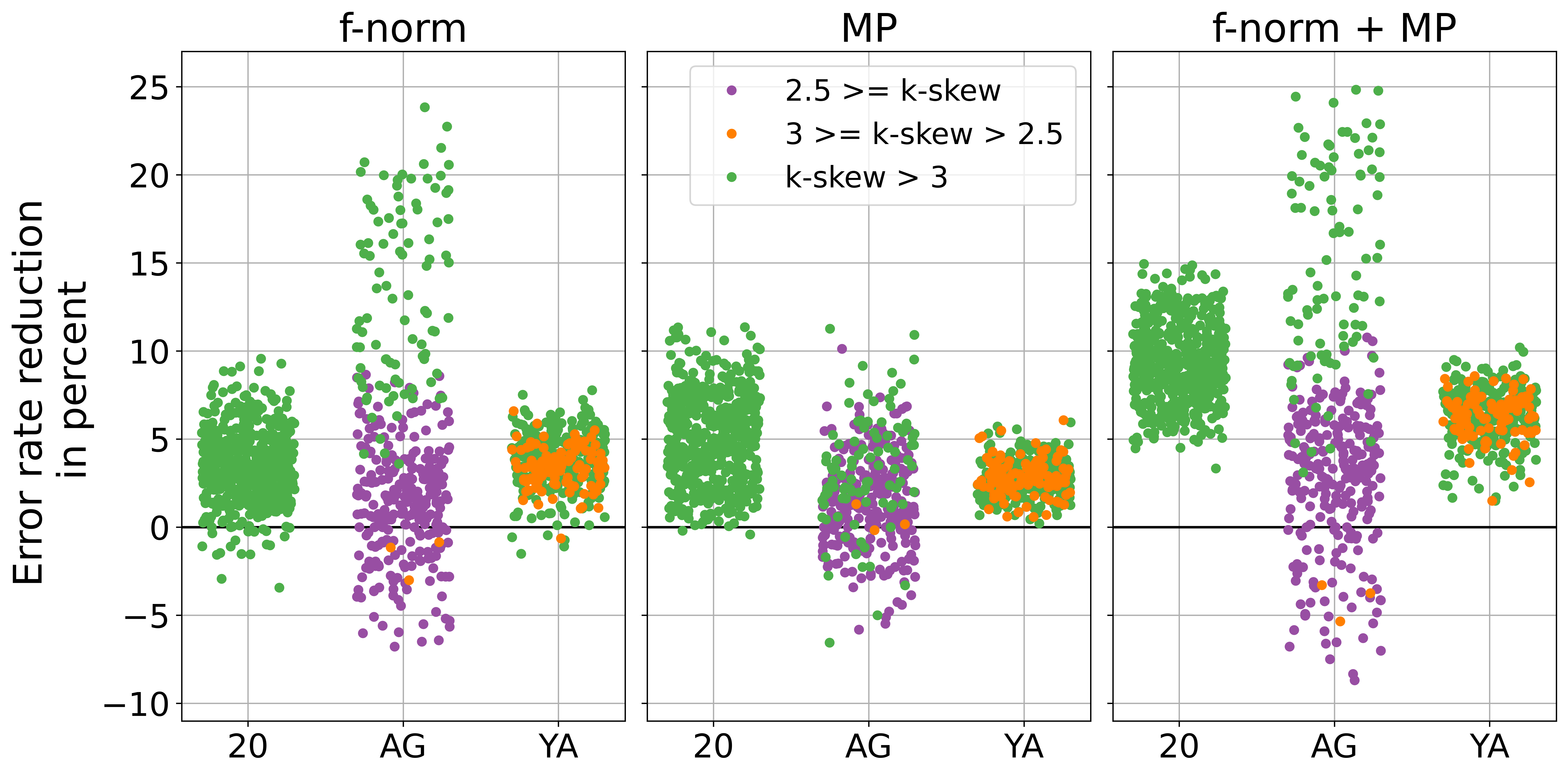}
\caption{\label{fig:error_rate_reduction}Error rate reduction for our models when using non-parametric standardization of the histogram (f-norm), mutual proximity (MP) as introduced by \cite{schnitzer2012local}, and the combination of f-norm + MP. Note that the base error rate varies for the models. This figure only shows the changes. If we only consider models where base embeddings have a k-skewness of more than 3, f-norm + MP reduces error rate in all cases. If we only consider embeddings where k-skewness is more than 2.5, f-norm + MP reduces error rate in all except 3 cases. Datasets: 20 Newsgroups (20), AG News (AG), 10\% of Yahoo Answers (YA).}
\end{figure*}

In this paper, we consider differently trained Sentence-BERT models and explore the effect on hubness scores and error rates of K-Nearest Neighbours (knn) classification when using post hoc hubness reduction methods. The overall picture is that the methods do reduce hubness and in most cases also reduce error rate, i.e., increase neighbor semantic similarity. For example, for a medium sized base model trained using cosine similarity and no normalisation we get a reduction of mean error rate over 12 seeds between $7\%$ and $13\%$ on the three used datasets. We also consider four pretrained models, trained on more data than our own and show that for the 20 Newsgroups data set, we can reduce hubness between $69\%$ and $83\%$ and reduce error rates between $7\%$ and $9\%$, i.e. training on more data does not solve the hubness problem. 

Our main contributions are:
\begin{itemize}
    \item Empirical evidence that high hubness can occur in Sentence-BERT embeddings, even when the model has been trained on large amounts of data.  
    \item A new use for a parameter free transformation of data (f-norm) as a hubness reduction method. This method forces the dimensions of embeddings to follow a standard normal distribution and has been used before by \citep{abrahamsen2011cure}, but never to our knowledge for hubness reduction.
    \item Experimental results which show the benefits of three hubness reduction methods f-norm, Mutual Proximity (MP) \citep{schnitzer2012local} and f-norm + MP and identify f-norm + MP as the best method for hubness and error rate reduction on three datasets. See figure \ref{fig:error_rate_reduction} for error rate reduction and figures \ref{fig:newsgroups_k_occurrence_distilroberta_euc_n_seed1}, \ref{fig:hubness_reductions} for hubness reduction.
\end{itemize}

\begin{figure*}[htb]
\includegraphics[width=\textwidth]{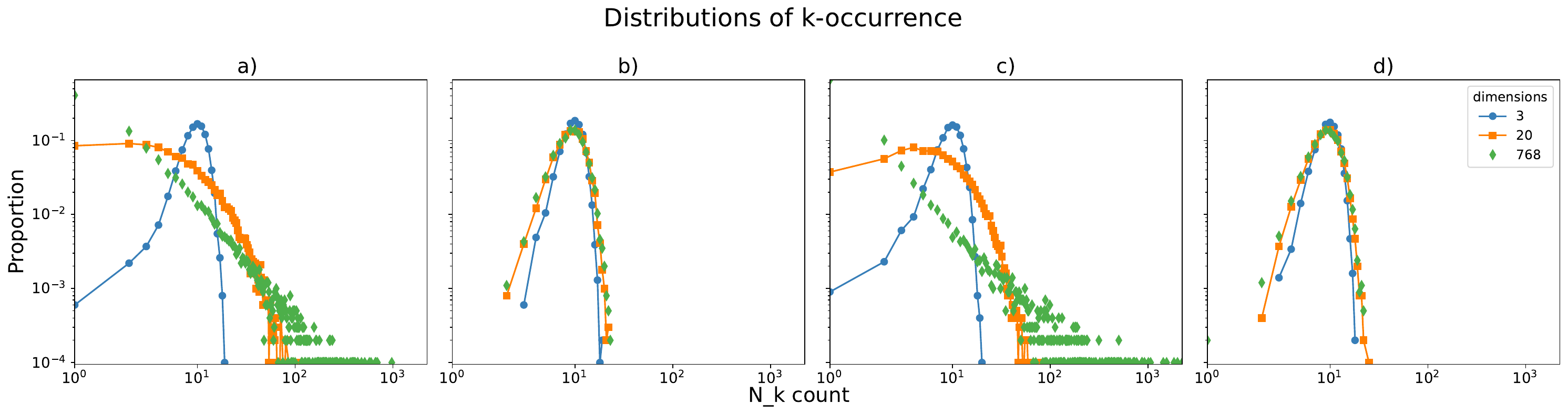}
\caption{\label{fig:std_normal_no_norm_norm_f_dist_no_norm_norm} Using 10 neighbours. Data with dimensions a) standard normally distributed (k-skewness: $-0.10$, $2.32$ and $11.62$,  rh: $0.09$, $0.35$ and $0.61$), b) standard normally distributed with embeddings normalized to unit length (k-skewness: $0.05$, $0.27$ and $0.37$,  rh: $0.08$, $0.11$ and $0.12$), c) F distributed (k-skewness: $-0.12$, $1.78$ and $19.30$, rh: $0.10$, $0.28$ and $0.74$) and d) F distributed after using f-norm (k-skewness: $0.04$, $0.34$ and $0.34$, rh: $0.9$, $0.11$ and $0.12$). b and d seem stable to high dimension.}
\end{figure*}

\subsection{Related Work}
In Radovanovic et al.\ \citep{radovanovic2009nearest} and \citep{radovanovic2010hubs}, they described why hubs appear in high dimensional data and illustrated the hubness problem on both synthetic and real data. \citep{feldbauer2019comprehensive} reviewed hubness reduction methods for a large variety of datasets. They concluded that e.g. local scaling \citep{zelnik2004self} and mutual proximity \citep{schnitzer2012local} formed state of the art in hubness reduction. Both \citep{radovanovic2010hubs} and \citep{feldbauer2019comprehensive} included text datasets in their analyses. However, both works are based on very high dimensional and sparse bag-of-word-representations. Hubness has also been found in word embeddings e.g.\ \citep{ormazabal-etal-2019-analyzing}. \citep{lazaridou-etal-2015-hubness} used hubness reduction to improve performance in zero-shot translation and image labelling. 

Recently, \citep{bogolin2022cross} introduced query bank normalisation to mitigate the hubness problem when using cross modal embeddings to retrieve images. Z-score normalisation has been used to reduce hubness of image embeddings \citep{fei2021z} and improve performance for few-shot learning. \citep{li-etal-2020-sentence} used the average of the context embeddings in the last layer of BERT as sentence embeddings, and showed that when using cosine similarity, accuracy of semantic similarity on text could be improved by transforming the distribution of the embeddings to a smooth isotropic normal distribution using normalizing flows. This idea is similar to our idea of making dimensions standard normally distributed, however, our method does not require training. \citep{timkey-van-schijndel-2021-bark} consider the problem of "rogue dimensions", a few dimensions of an embedding which are far from the origin, have high variance and dominate the cosine similarity, so that these few dimensions decide whether two embeddings are similar or not, even though they do not have much say in model decisions. See section \ref{sec:discussion} for further relations to our work.   %

\section{Hubness}
\label{sec:analysis}
\label{sec:emergence}
Let $ X = (x_1, ..., x_m) \subset \mathbb{R}^D$ be $m$ points in $D$-dimensional space. We define $N_k(x)$, called the $k$-occurrence, for a point $x \in X$ as the number of times $x$ occurs in the $k$ nearest neighbours of all points in $X$, for some distance measure. In this paper we will use Euclidean distance. \label{sec:quantify}
We will use the k-occurrence, $N_k(x)$, to measure the hubness of our embedding spaces. However, for comparing several types of embeddings it is more convenient with a single score, so we will use the skewness of the $N_k(x)$ distribution ($k$-skewness) as proposed in \cite{radovanovic2010hubs} and the robinhood score (rh) as proposed in \cite{feldbauer2018fast}. The robinhood score, $H^k$, is given by 
\begin{equation}
    H^k = \frac{\sum_{x \in X}\lvert N_k(x) - k \rvert}{2k(m - 1)}
\end{equation}
and measures how far the $k$-occurrences of the embeddings are from the expected $k$. That is, it is a fraction between 0 and 1, which tells us how large a part of nearest neighbours needs to be changed if all objects are to be in the k nearest neighbours of the same number of points. We use the scikit-hubness library by \cite{feldbauer2020scikithub} to calculate these scores. In this paper, we use $k = 10$ for measuring $k$-skewness and robinhood score, since the exact value is not important and $10$ is the value used in much of the literature we reference. 

\begin{algorithm}
\caption{f-norm}\label{alg:f-norm}
\begin{algorithmic}
\State $X = \{x_{i, d}\}_{i \in \{1, ..., m\}, d \in \{1, ..., D\}}$
\For{$d \in \{1, ..., D\}$} 
\State $Z \gets (z_1, ..., z_m) \sim \mathcal{N}(0, 1)$
\State $idxData \gets \text{argsort}_{i \in \{1, ..., m\}}(x_{i, d})$
\State $idxNormal \gets \text{argsort}_{i \in \{1, ..., m\}}(z_{i})$
\For{$idx_1, idx_2 \in (idxData, idxNormal)$} 
\State $x_{idx_1, d} \gets z_{idx_2}$
\EndFor
\EndFor
\For{$i \in \{1, ..., m\}$} 
\State $x_{i} \gets \frac{x_{i}}{\Vert x_{i} \Vert_2  }$
\EndFor
\end{algorithmic}
\end{algorithm}

\begin{figure*}
\includegraphics[width=\linewidth]{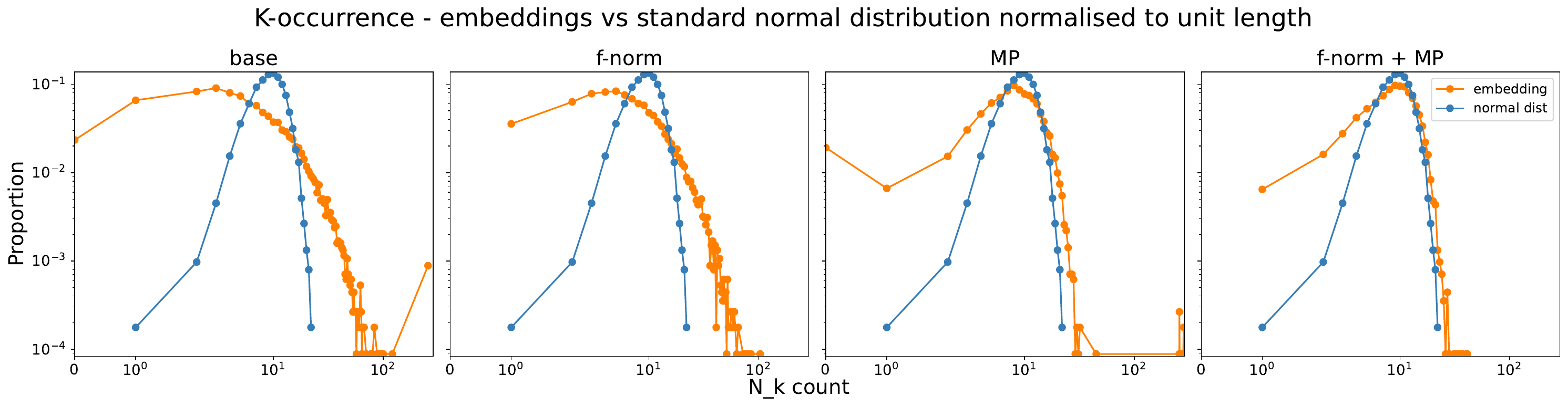}
\caption{\label{fig:newsgroups_k_occurrence_distilroberta_euc_n_seed1}$K$-occurrence distribution (10 neighbours) for embeddings on 20 newsgroups (orange) compared to points coming from a standard normal distribution in same number of dimensions (blue) with embeddings normalised to unit length. Using our reduction methods, the $k$-occurrence of the embeddings come closer to the stable distribution. Note MP still leaves some large hubs. Embedding $k$-skewness: $8.79$, $2.20$, $20.98$, $0.42$, rh: $0.35$, $0.30$, $0.20$, $0.16$. Model: distilroberta-base\_euclidean\_n\_seed1.}

\end{figure*}

According to \cite{radovanovic2010hubs}, in unimodal data distributions there is a strong correlation between $N_k(x)$ and how close $x$ is to the global data mean. That is, points close to the center of the data tend to become hubs. Therefore, we expect hubness to be lower when all points have equal distance to the mean, which will be the case if for example all points are on the surface of a hypersphere (more in appendix \ref{k-occurrence_data_mean}). Which means that if our data is standard normally distributed in each dimension, and we then normalise to unit length, we should get lower hubness. In figure \ref{fig:std_normal_no_norm_norm_f_dist_no_norm_norm}a) and b) we see that this is indeed the case. In LLM derived embeddings we do not expect the individual dimensions to be distributed according to the same normal distribution, however, we find empirically that the distributions of the dimensions of sentence BERT embeddings do look close to normal distributions (examples in Appendix figure \ref{fig:emb_dim_examples}). See more motivation in Appendix \ref{sec:synthetic}. 
Thus, we explored the idea to force the dimensions of the data points to follow a standard normal distribution, but preserving the ranking of the values (we call this method `f-norm', Algorithm \ref{alg:f-norm}).
Note, when using this method the distribution of the dimensions before the transformation no longer matters. See figure \ref{fig:std_normal_no_norm_norm_f_dist_no_norm_norm}c) and d) for an example with F distributed dimensions. See figure \ref{fig:newsgroups_k_occurrence_distilroberta_euc_n_seed1}  and appendix figure \ref{fig:spy_before_after} for how f-norm affects some real data. 

\subsection{Methods for hubness reduction of semantic spaces}
\label{sec:methods}
Normalisations used during training aim to center the data and/or make it lie on the unit sphere, see Appendix \ref{sec:synthetic}. In our models we used: No normalisation ('none'), normalisation of embeddings to unit length ('n'), centering of embeddings (subtracting the mean value of the embedding from all dimensions in the embedding) ('c'), a combination of centering and normalisation of embeddings to unit length ('c,n') and z-score normalisation of embeddings (according to \cite{fei2021z} this had an effect when doing few shot learning) ('z').

After training and obtaining embeddings for a dataset using a model, we tested three different post hoc methods for hubness reduction on the three datasets. 
\begin{itemize}
\item Forcing standard normal distribution in each dimension of embeddings and then normalising embeddings to unit length (f-norm) as described in Algorithm \ref{alg:f-norm}.
\item Mutual proximity (MP) as introduced by \cite{schnitzer2012local}.
\item First forcing a standard normal distribution in each dimension and normalising embeddings, then using mutual proximity on the result (f-norm + MP).
\end{itemize}
On the 20 Newsgroups dataset, we tested two additional methods: f-uniform and local scaling. The latter was introduced by \cite{zelnik2004self} and proposed for hubness reduction by \cite{schnitzer2012local}. More in Appendix \ref{additional_hubness_reduction_methods}. 

MP aims to make the nearest neighbour relation more symmetric by translating the distance between points $d(x, y)$ to a probability $MP(d_{x,y})$ that $x$ is the nearest neighbour of $y$ and that $y$ is also a nearest neighbour of $x$ in the following way: Let $p(d_x)$ be the probability density function of the probability distribution of distances from $x$ to all other points in the dataset. Then use $p(d_x)$ and $d(x, y)$ to find the probability of $y$ being the nearest neighbour of $x$. Then use $p(d_y)$ and $d(x, y)$ to find the probability of $x$ being the nearest neighbour of $y$. It is then these probabilities which are used as the distance in the knn algorithm. \citep{schnitzer2012local} showed that it is enough to calculate distances to a sample of other points. More in Appendix \ref{sec:mp_implementation}.

\section{Evaluation}
\label{sec:evaluation}

We trained Sentence-BERT models using three different base models, which were not trained on the datasets we use for testing,  (microsoft-MiniLM-L12-H384-uncased\footnote{\url{huggingface.co/MiniLM-L12-H384-uncased}} \cite{wang2020minilm}, distilroberta-base\footnote{\url{huggingface.co/distilroberta-base}} \cite{sanh2019distilbert},  microsoft-mpnet-base\footnote{\url{huggingface.co/microsoft/mpnet-base}} \cite{song2020mpnet}) and three different distance measures during training (cosine similarity ('cos'), cosine distance ('cos\_dist'), Euclidean distance ('euclidean')). For each combination of model, distance measure and normalisation method from \ref{sec:methods}, we trained with twelve random seeds (model naming scheme is [base model]\_[distance measure]\_[normalisation]\_[seed]). This gave us $540$ models in all. On the large datasets (AG News and Yahoo Answers), we only calculated embeddings and results for models using the medium and large base models, that is distilroberta-base and microsoft-mpnet-base, so on these datasets we have results for $360$  models. We trained our models on the STS benchmark dataset\footnote{\url{sbert.net/datasets/stsbenchmark.tsv.gz}} for $4$ epochs. 
More in Appendix \ref{sec:comp_details}, \ref{sec:versions}, \ref{sec:sts_details}. \par 

We also made embeddings using four pretrained Sentence-BERT models, which have been trained on more data than our own. These models were: all-mpnet-base-v2\footnote{\url{huggingface.co/sentence-transformers/all-mpnet-base-v2}}, multi-qa-distilbert-cos-v1\footnote{\url{huggingface.co/sentence-transformers/multi-qa-distilbert-cos-v1}}, all-distilroberta-v1\footnote{\url{huggingface.co/sentence-transformers/all-distilroberta-v1}} and all-MiniLM-L12-v2\footnote{\url{huggingface.co/sentence-transformers/all-MiniLM-L12-v2}}. Note that all these pretrained models have been trained on Yahoo Answers in various ways. More in Appendix \ref{sec:pretrained_models}.  

We used three text classification datasets: 20 Newsgroups, AG News (by Antonio Gulli) and ten percent of the Yahoo Answers dataset \cite{zhang2015character}. Details in Appendix \ref{sec:app_datasets}.  

We classify using knn. To identify the number of neighbours to use in knn for classification, we used stratified 10-fold validation on the training set. We measured error rate on the test set. More details in Appendix \ref{sec:eval_details}. 

Let the train embeddings be an $m\times d$ matrix and the test embeddings be an $l\times d$ matrix. When using f-norm at test time, we append the test embeddings to the train embeddings matrix giving us an $(m + l) \times d$ matrix, which we then use the algorithm on.

\section{Results}
\label{sec:results}
\begin{figure}
\includegraphics[width=0.49\textwidth]{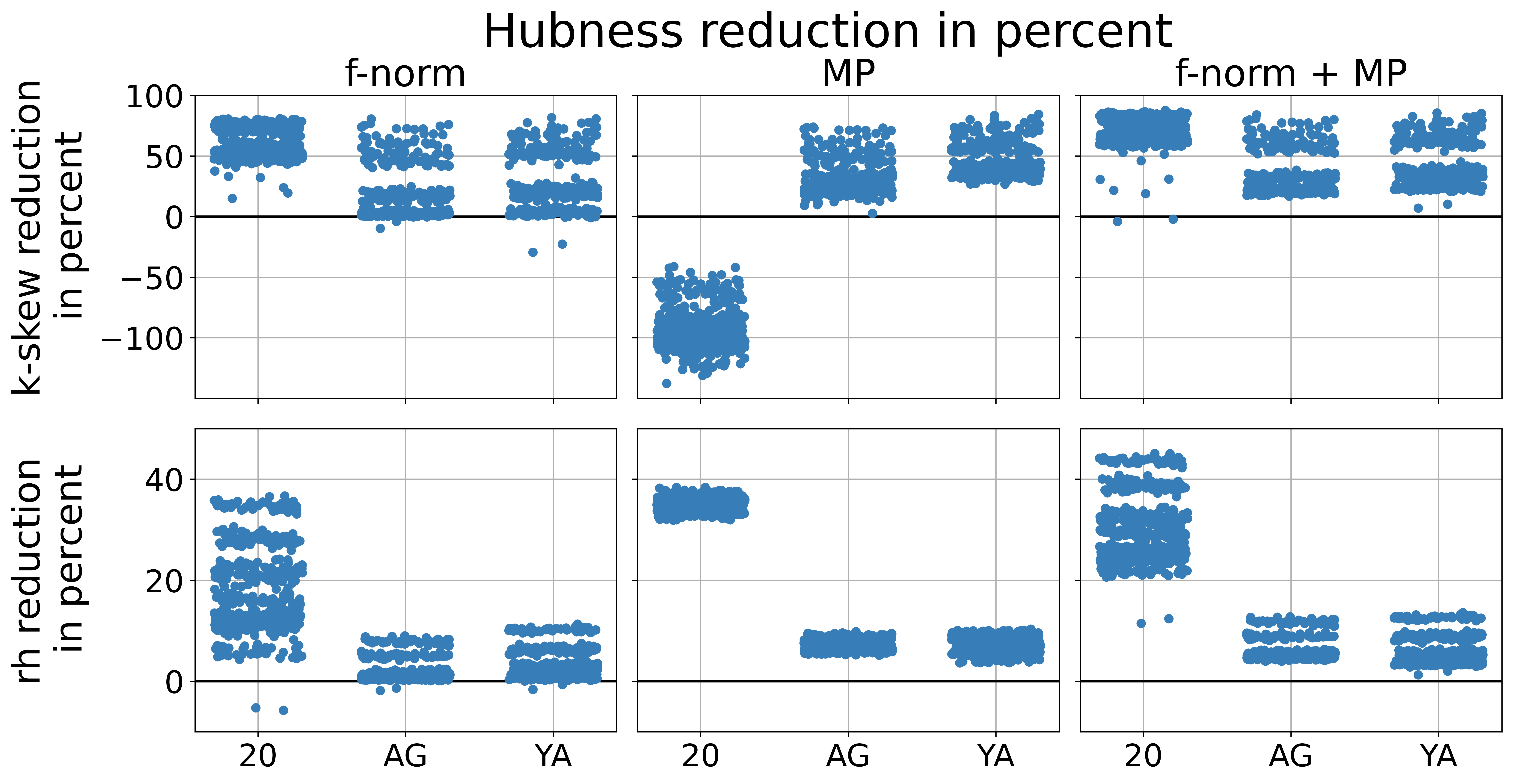}
\caption{\label{fig:hubness_reductions}Reduction in hubness using f-norm, MP and f-norm + MP measured with $k$-skewness ($k$-skew) and robinhood score (rh). Higher is better. Hubness is reduced in almost all cases, however for MP on 20 Newsgroups hubness increases as measured with $k$-skewness. Likely induced by the hubs remaining after MP, see figure \ref{fig:newsgroups_k_occurrence_distilroberta_euc_n_seed1}.}
\end{figure}

Trained, models, embeddings and raw result files can be found at DTU Data\footnote{\url{https://doi.org/10.11583/DTU.c.6165561.v1}}. Code can be found on GitHub\footnote{\url{https://github.com/bemigini/hubness-reduction-improves-sbert-semantic-spaces}}. Overviews of training combinations and post hoc reduction methods with error rate and hubness scores are in Appendix \ref{sec:overviews}. See this material for results on f-uniform and local scaling. Summary of error rate changes from the overview tables is in figure \ref{fig:error_rate_reduction}, summary of the changes in hubness is in figure \ref{fig:hubness_reductions}. 

We see that the methods reduce hubness as measured with the robinhood score in almost all cases (figure \ref{fig:hubness_reductions}). However, when using MP, $k$-skewness went up in all cases on 20 Newsgroups. Inspecting the $k$-occurrence distribution, this seems to be because while MP makes the knn relation more symmetric for most data points, there are still some rather large hubs which it does not change. These hubs are sentences which are either empty or which only contain whitespace characters. In 20 Newsgroups there are 268 sentences which are either empty or whitespace. Before hubness reduction, 252 of them have a $k$-occurrence of 0 and 10 have a $k$-occurrence of more than 100. After using MP, 191 of them have a $k$-occurrence of 0 and 10 still have a $k$-occurrence of more than 100. However, when using f-norm + MP there are no longer these very large hubs (figure \ref{fig:newsgroups_k_occurrence_distilroberta_euc_n_seed1}), in fact after f-norm + MP, all the empty or whitespace sentences have k-occurrences between 4 and 17. Thus, these sentences are no longer disconnected with respect to the neighbourhood relation. Using f-norm + MP also decreases k-skewness in more cases than using just f-norm.

\begin{figure}
\includegraphics[width=0.49\textwidth]{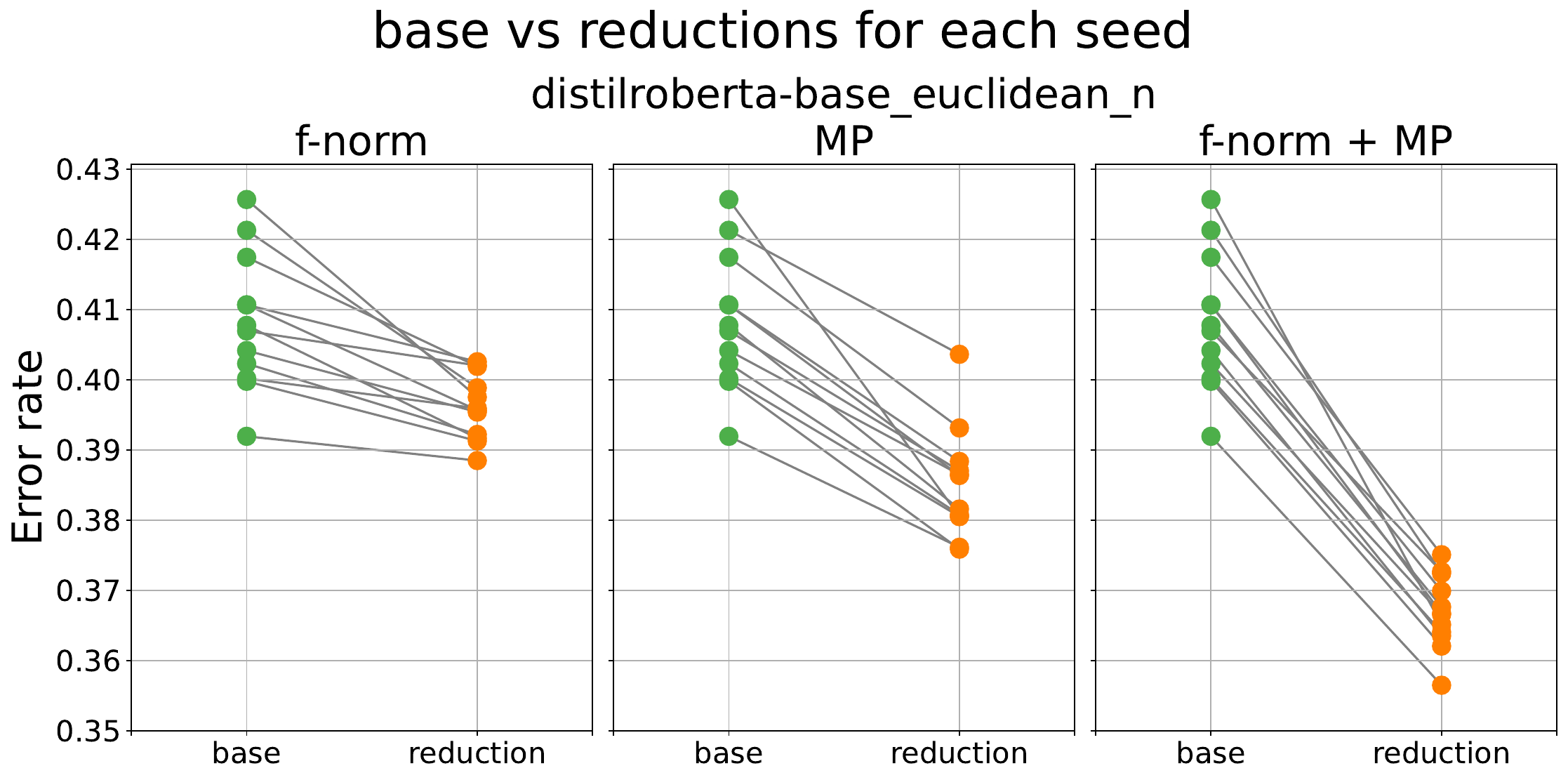}
        \caption{knn error on 20 Newsgroups before vs after hubness reductions. Average base $k$-skewness $6.06$. See \ref{sec:evaluation}}\label{fig:base_vs_reductions_distilroberta-base_euclidean}
\end{figure}

On all three datasets, when $k$-skewness is more than $3$, error rate is reduced when using f-norm + MP (figure \ref{fig:error_rate_reduction}). See figure \ref{fig:base_vs_reductions_distilroberta-base_euclidean} for examples of absolute error reduction. More examples in Appendix \ref{sec:error_base_vs_reductions}. Embeddings on AG News had lower base hubness and error rate than embeddings on the other two datasets, however, since f-norm reduces error rate for all embeddings with $k$-skewness more than 3, we think it likely that the low hubness makes it difficult for the hubness reduction methods to improve classification performance on this dataset. We also found that f-norm + MP is more likely to have a larger effect on error rate, if the base skewness is higher (Appendix \ref{sec:more_skew_more_effect}). We hypothesize that when hubness is not high enough, the transformation of the embeddings harms the learned structure of the embeddings more than it helps.

The lowest mean error rates came from models using the largest base model. However, using f-norm + MP can in some cases make up for the model size. For example, distilroberta-base\_cos\_none base and microsoft-mpnet-base\_cos\_none had base mean error of $10.9$\% and $9.4$\% on AG News. Using f-norm + MP reduces the error to $8.8$\% and $8.4$\%. So the smaller model with f-norm + MP is better than the bigger model without. More about the smallest error rates in Appendix \ref{sec:best_error_rate_details}. 

We found that models which use some kind of embedding length normalisation (n; c,n; z) tend to have slightly lower hubness measured with robinhood score than models which do not have this kind of normalisation (none; c). More on differences between normalisations during training in Appendix \ref{sec:normalisation_during_training_differences}. 
We found that which distance measure is used during training is not a good predictor for error rate or effectiveness of hubness reduction. On AG News, the ten best mean error rates all came from models which used Euclidean distance for training, however on 20 Newsgroups, models trained with Euclidean distance had worse error rate than models trained with cosine similarity or cosine distance.

\begin{table*}[ht]
  \caption{\label{table:pretrained_error_skew} Error rate (error) and k-skewness (skew) on 20 Newsgroups (20), AG News (AG) and ten percent of Yahoo Answers (YA) for four pretrained models, base case vs after hubness reduction. Note that all these models have been trained on YA in various ways and their hubness on YA is in general lower than the models we trained. Thus, it seems that training on a specific dataset can make hubness go down for that dataset. However, hubness is still high on 20 which the models have not been trained on.}
  \centering
  \begin{tabular}{l l | c c | c c | c c }
    \toprule
    \multicolumn{2}{c |}{\textbf{Model}} & \multicolumn{2}{c |}{\textbf{20}} & \multicolumn{2}{c |}{\textbf{AG}} & \multicolumn{2}{c}{\textbf{YA}}          \\
     & & \textbf{error} & \textbf{skew} & \textbf{error} &  \textbf{skew} & \textbf{error} & \textbf{skew}  \\
    \midrule
    
\textbf{multi-qa-distilbert-cos-v1} & \textbf{base}         & 0.318 & 7.58  & 0.081 & 1.90 & 0.271 & 2.12  \\ 
                                    & \textbf{f-norm + MP} & 0.289 & 1.89  & 0.079 & 1.57 & 0.269 & 1.88  \\
\midrule
\textbf{all-MiniLM-L12-v2} & \textbf{base}        & 0.324 & 11.45 & 0.088 & 1.91 & 0.268 & 1.98 \\
                           & \textbf{f-norm + MP} & 0.300 & 1.88  & 0.085 & 1.56 & 0.267 & 1.79 \\
\midrule
\textbf{all-mpnet-base-v2} & \textbf{base}       & 0.295 & 11.53 & 0.082 & 1.81 & 0.251 & 2.02  \\
                           & \textbf{f-norm + MP} & 0.272 &  1.98 & 0.082 & 1.54 & 0.250 & 1.78  \\
\midrule
\textbf{all-distilroberta-v1} & \textbf{base}          & 0.322 &  6.61 & 0.085 & 1.93 & 0.254 & 2.08  \\
                              & \textbf{f-norm + MP} & 0.296 &  2.05 & 0.085 & 1.59 & 0.253 & 1.82  \\
    \bottomrule
  \end{tabular}
\end{table*}

\subsection{Pretrained Models}
All the tested pretrained models have a high base hubness as measured with $k$-skewness on 20 Newsgroups (See Table \ref{table:pretrained_error_skew}). This means that training models on more data will not necessarily result in low hubness. On the 20 Newsgroup dataset, the error rate goes down (reduction between $7\%$ and $9\%$.) when using f-norm + MP for all four models. These reductions are highly significant, since using McNemar's test gives p-values smaller than $2.1 \cdot 10^{-9}$. However, error rate is not reduced for the other two datasets. We interpret this as being due to the low base hubness of the pretrained models on AG News and Yahoo Answers. AG News also had a low base hubness for most of our own models, so we suppose this is due to the nature of the dataset. Our own models had higher base hubness on Yahoo Answers than the pretrained models, however all the pretrained models were also trained on Yahoo Answers in various ways. Therefore, we assume that training on a specific dataset can make hubness go down for that dataset. See Table \ref{error_rate_hub_pretrained} in Appendix \ref{sec:pretrained_models} for more details.

\section{Discussion and Conclusion}
\label{sec:discussion}

Since f-norm also reduces the rogue dimensions problem \cite{timkey-van-schijndel-2021-bark}, the reductions in error might not be solely due to reduced hubness. However, in cases where MP reduces error rate, we hypothesize that hubness reduction does contribute to the error reduction. \par 
We measured hubness with both $k$-skewness and the robinhood score. The robinhood score is less sensitive to outliers, so it can detect when a method reduces the overall hubness even though it leaves some outliers, while $k$-skewness might grow in such cases. However, $k$-skewness is a better predictor for whether hubness reduction methods using f-norm will have a positive effect on the error rate. The robinhood score has a different baseline scale for each dataset, hampering its interpretation. \par
We have only looked at how hubness reduction methods affect knn. Since we used Euclidean distance in our knn, the error rates on our datasets tell us how well the Euclidean distances between datapoints correspond to class structure. Thus, when searching for other methods which might benefit from these hubness reduction methods, we should look for models which use distances between the datapoints for classification. Other clustering or classification techniques might not be improved using these methods (Note on SVM in Appendix \ref{sec:svm_note}).
When using f-norm for $m$ train embeddings and $l$ test embeddings of dimension $d$, one will have to sort $(m+l) \cdot 2d$ numbers. However, for very large training datasets, it should not be necessary to use all the training data. It would be an interesting direction for further research to find out how many and which training points to keep to get good performance. 
 
In conclusion, when using neighbourhood based approaches and high dimensional text embeddings, one should always check for hubness. For a  broad set of training choices, if there is high hubness in embeddings (k-skewness of more than $3$ seems to be useful as a rule of thumb), then hubness reduction techniques (especially f-norm + MP) will produce better text representations as shown by more symmetric nearest neighbour relation and improved classification.

\section*{Acknowledgements}
This work was supported by the Danish Pioneer Centre for AI, DNRF grant number P1 and by the Novo Nordisk
Foundation grant NNF22OC0076907 ”Cognitive spaces - Next generation explainability”.
Thanks to Mikkel N. Schmidt for helpful comments on the abstract and introduction.

\printbibliography

\clearpage

\appendix

\addcontentsline{toc}{section}{Appendix} 
\part{Appendix} 
\parttoc 

\section{Datasets}
\label{sec:app_datasets}
We used text classification datasets, where we expect the classes to be related to the semantic meaning of the texts. 
\begin{itemize}
    \item 20 Newsgroups as can be fetched through scikit-learn which fetches data from the 20 newsgroups website\footnote{\url{http://qwone.com/~jason/20Newsgroups/}}. Originally collected by Ken Lang. It consists of 18846 newsgroup posts with 20 different topics split into a train set of 11314 samples and a test set of 7532 samples. Embeddings were made on the posts as given by the scikit-learn library with no additional preprocessing. License: Unknown. 
    
    \item AG News fetched from Kaggle\footnote{\url{www.kaggle.com/datasets/amananandrai/ag-news-classification-dataset}}. Originally made by Antonio Gulli \footnote{\url{groups.di.unipi.it/~gulli/AG_corpus_of_news_articles.html}}. News stories divided into four classes each with 30,000 training samples and 1,900 testing samples. So in all we have 120,000 training samples and 7,600 test samples. Embeddings were made on the title concatenated with the description with line breaks replaced by spaces. License: Custom. Any non-commercial use allowed.
    \item Yahoo Answers fetched from Kaggle\footnote{\url{www.kaggle.com/datasets/yacharki/yahoo-answers-10-categories-for-nlp-csv}} made by \cite{zhang2015character}. Questions and best answers from Yahoo Answers from 10 categories. We use 10 percent of this dataset, so for each category we have 14,000 training samples and 600 test samples, which means we have 140,000 training samples and 6000 test samples in all. Indexes of used samples can be found in the code repository\footnote{\url{https://github.com/bemigini/hubness-reduction-sentence-bert}}. Embeddings were made on the question title concatenated with the question content and the best answer with line breaks replaced by spaces. License: Unknown.    
\end{itemize}

\section{K-occurrence and the Mean of the Data}
\label{k-occurrence_data_mean}
According to \cite{radovanovic2010hubs}, in unimodal data distributions there is a strong correlation between $N_k(x)$ and how close $x$ is to the global data mean. That is, points close to the center of the data tend to become hubs. While for multimodal distributions \cite{radovanovic2010hubs} found that points close to local cluster means tend to become hubs. \citep{radovanovic2010hubs} also argue that hubness depends on the intrinsic dimension of data and not the dimension of the embedding. \cite{low2013hubness} points out that high intrinsic dimension does not in itself produce high hubness. They argue that high density gradients, e.g., introduced by support boundaries can lead to formation of hubs.

\section{Additional Hubness Reduction Methods}
\label{additional_hubness_reduction_methods}
On the 20 Newsgroups dataset we tested two additional hubness reduction methods:
\begin{itemize}    
\item Forcing uniform distribution centered at 0 in each dimension of embeddings and then normalising embeddings to unit length (f-uniform) same way as described in section 2.1 "Illustration on Synthetic Data".
\item Local scaling as introduced by \cite{zelnik2004self} (Proposed for hubness reduction by \citealt{schnitzer2012local}).
\end{itemize}
We tested f-uniform, since it might not be the distribution which is important, but only that the center of the data is the zero vector and that the variance of the dimensions is the same. However, we did expect that f-uniform would destroy the learned structure of the embeddings, since it seems the distribution of the dimensions of the embeddings before hubness reduction is close to normal. We were suprised to find that f-uniform decreased error rate in almost as many cases as f-norm, but we still decided to only continue with f-norm on the two other datasets, because of the observed distribution of the embeddings.   \par 
Local scaling introduces a secondary distance measure $LS(d_{x,y})$ between points $x, y$, which uses a local scaling parameter $\sigma_{x}$ - the distance between $x$ and the m'th nearest neighbour of $x$ for a chosen $m \in \mathbb{N}$. 
\begin{equation}
    LS(d_{x,y}) = 1 - \exp\left(-\frac{d_{x,y}^2}{\sigma_{x}\sigma_{y}}\right)
\end{equation}
We used $m = 5$. \par 

We decided not to continue with local scaling on the two other datasets, since it did not decrease error rate in as many cases as the other hubness reduction methods. 

\section{Illustration on Synthetic Data}
\label{sec:synthetic}
Figure \ref{fig:std_normal_no_norm_norm_f_dist_no_norm_norm}a) shows the distribution of the $k$-occurrence for $10,000$ data points drawn from a standard normal distribution in $3$, $20$ and $768$ dimensions ($768$ is the dimensionality of Sentence-BERT embeddings). Recall that the density function for a $D$ dimensional standard normal distribution is
\begin{equation}
    (2\pi)^{-D/2}e^{-\frac{1}{2}\sum_{i = 1}^{D} x_i^2} = (2\pi)^{-D/2}e^{-\frac{1}{2}\lVert x \rVert^2}.
\end{equation}
So the distribution is a function of the length of the vector. Hence, by normalising data points to unit length, we obtain a uniform distribution on the hypersphere. Because of \cite{radovanovic2010hubs} we expect hubness to be lower when all points have equal distance to the mean, see appendix \ref{k-occurrence_data_mean}. Note, the distribution of the $k$-occurrence is symmetric with this normalization even in 768 dimensions, see figure \ref{fig:std_normal_no_norm_norm_f_dist_no_norm_norm}b), thus for normal distributed data normalization reduces hubness. 

Drawing data points from a normal distribution with a nonzero mean vector, makes the $k$-occurrence skewed as before even if we normalise embeddings to unit length. However, if we center the data before normalising, we get again a symmetric $k$-occurrence distribution. If we instead use a uniform distribution (e.g., the interval $[-1, 1]$) for each dimension and again normalise embeddings to unit length, then this $k$-occurrence will also become symmetric. 

However, in LLM derived embeddings we do not expect the individual dimensions to be distributed according to the same normal distribution although empirically, the distributions of the dimensions do look like normal distributions (examples in Appendix figure \ref{fig:emb_dim_examples}).

\section{Spyplot of before and after f-norm}
\begin{figure*}
\includegraphics[width=0.98\linewidth]{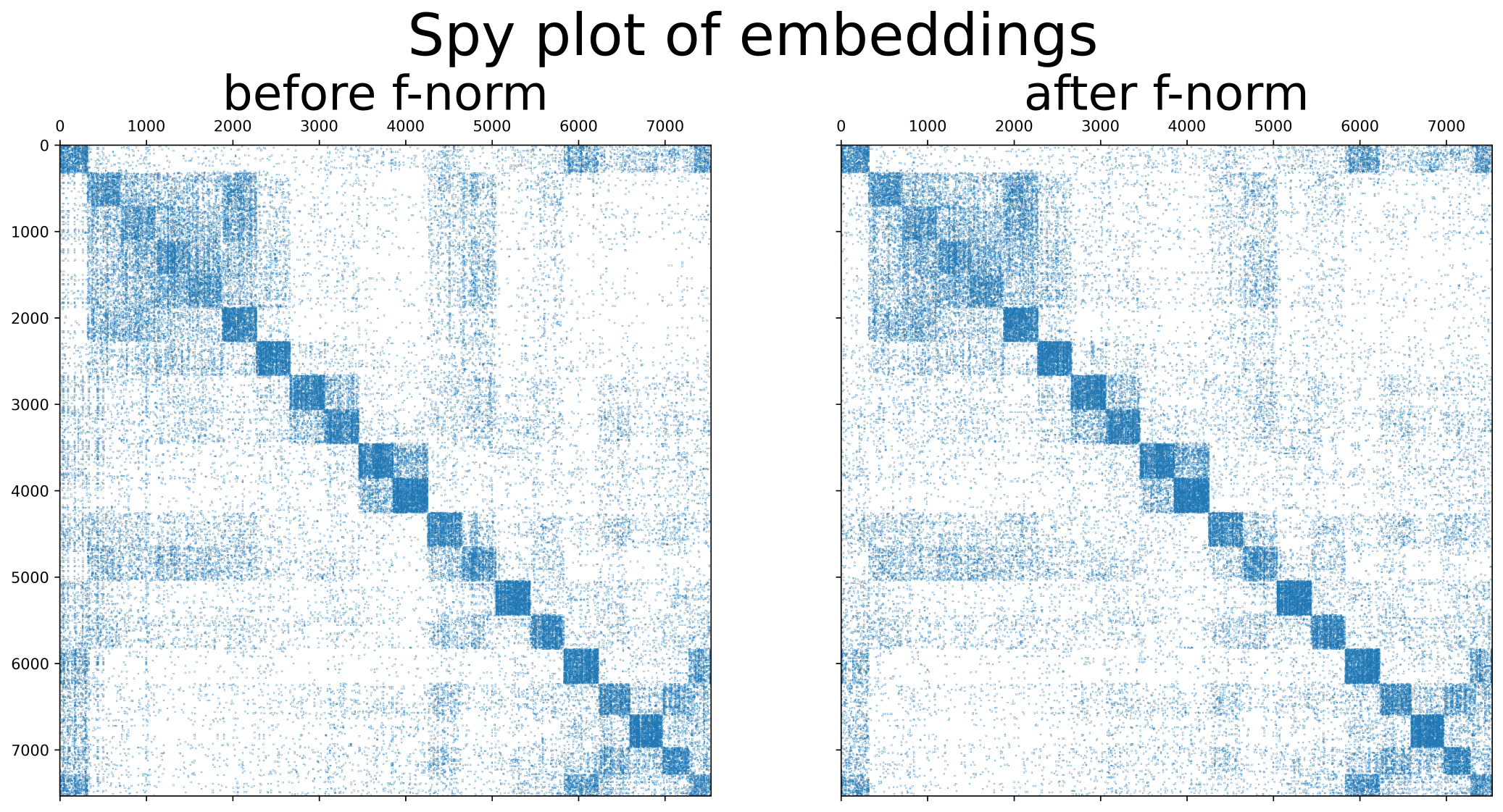}
\caption{\label{fig:spy_before_after} The model microsoft-MiniLM-L12-H384-uncased\_cos\_n with seed 1 on 20 Newsgroups test split. Column samples in the 10 nearest neighbours of the row sample are marked with a blue dot, otherwise they are unmarked. Samples are sorted by category. Hubs can be seen as vertical lines especially in the left side of the before plot. These become much less pronounced after having used f-norm.}
\end{figure*}

Figure \ref{fig:spy_before_after} shows how the number of other sentences a sentence is in the 10 nearest neighbours of is reduced by f-norm. 

\section{Example of Distribution of Dimensions}
\begin{figure}[ht]
  \includegraphics[width=0.49\textwidth]{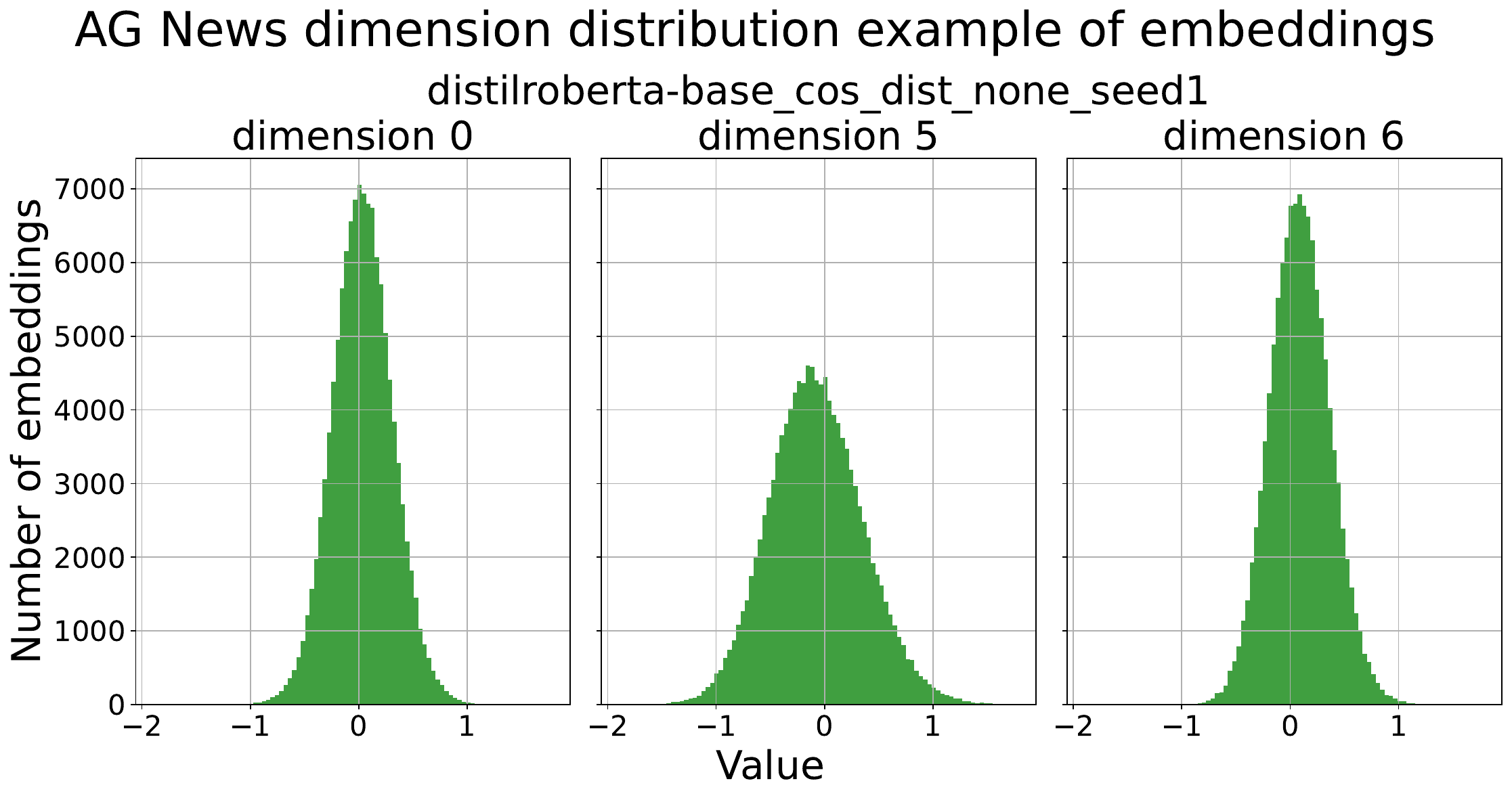}
        \caption{\label{fig:emb_dim_examples}Values in the dimensions are roughly normally distributed, but do not have the same mean or variance.}
\end{figure}

See figure \ref{fig:emb_dim_examples}.

\section{Error rate - Base vs reductions}
\label{sec:error_base_vs_reductions}
\begin{figure}[ht]
  \includegraphics[width=0.99\linewidth]{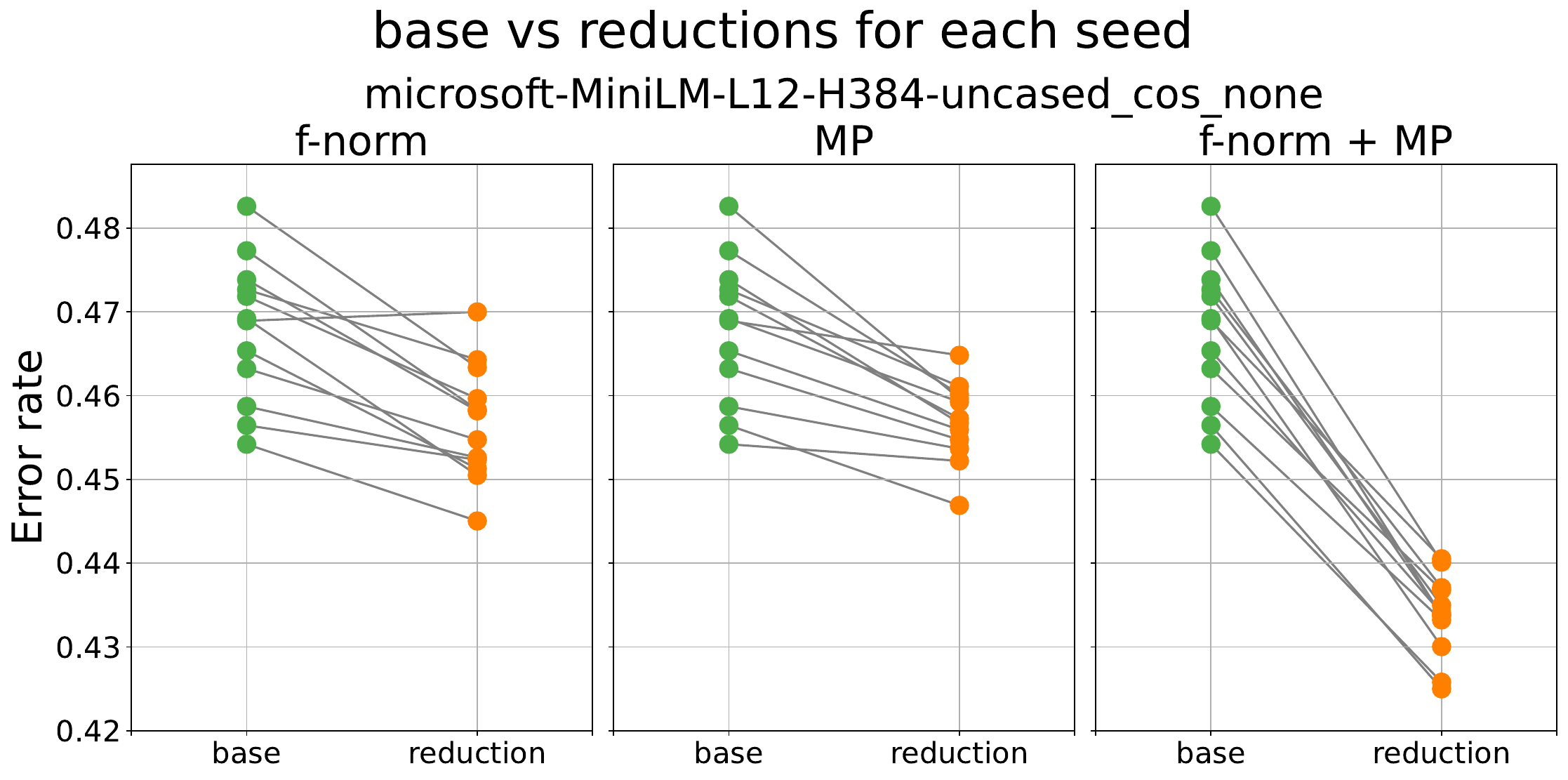}
\caption{20 Newsgroups. Average base k-skewness 8.15.}\label{fig:base_vs_reductions_microsoft-MiniLM-L12-H384-uncased_cos}
\end{figure}

\begin{figure}[ht]
  \includegraphics[width=0.99\linewidth]{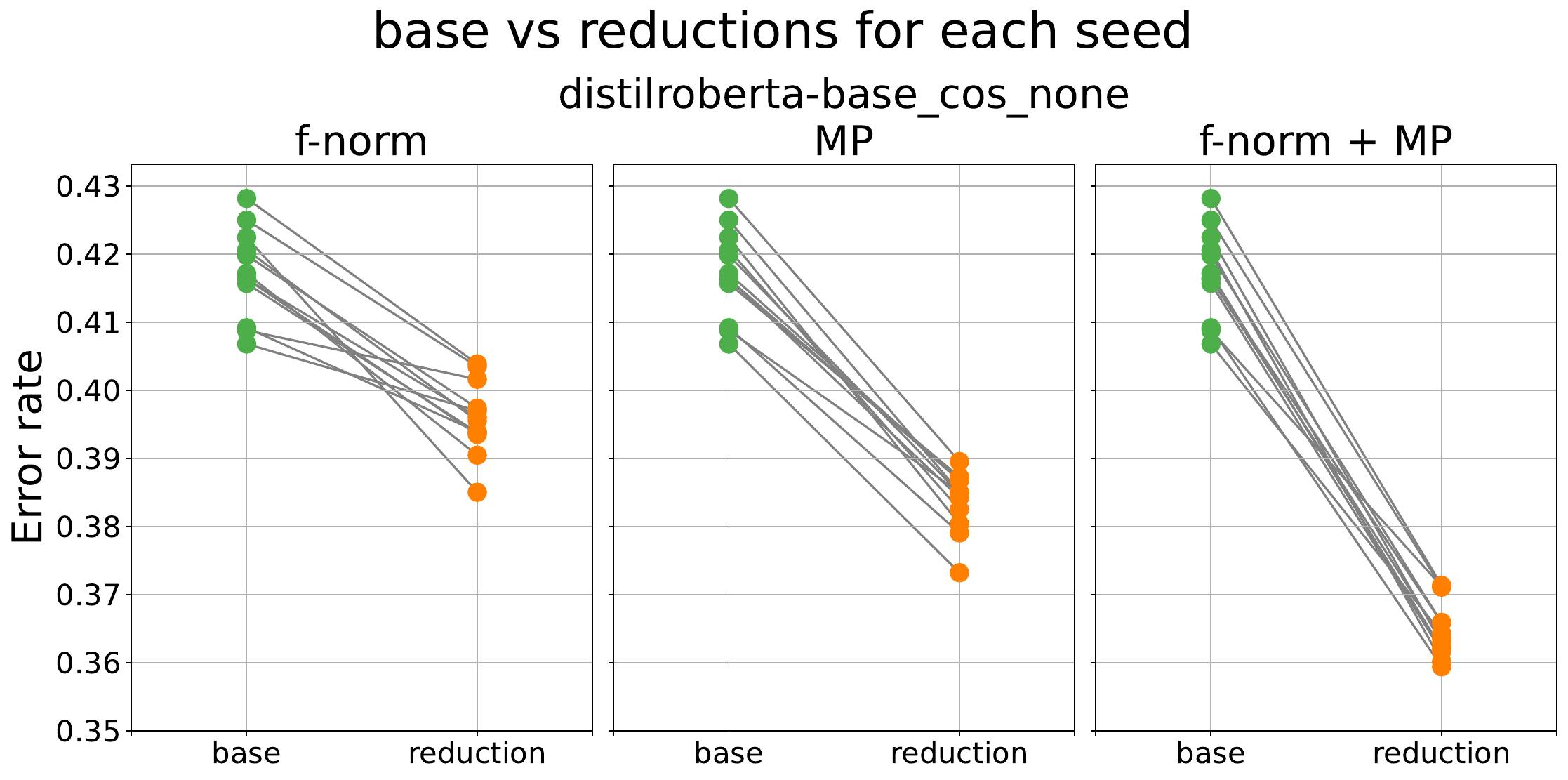}
\caption{20 Newsgroups. Average base k-skewness 5.60.}\label{fig:base_vs_reductions_distilroberta-base_cos}
\end{figure}

\begin{figure}[ht]
  \includegraphics[width=0.99\linewidth]{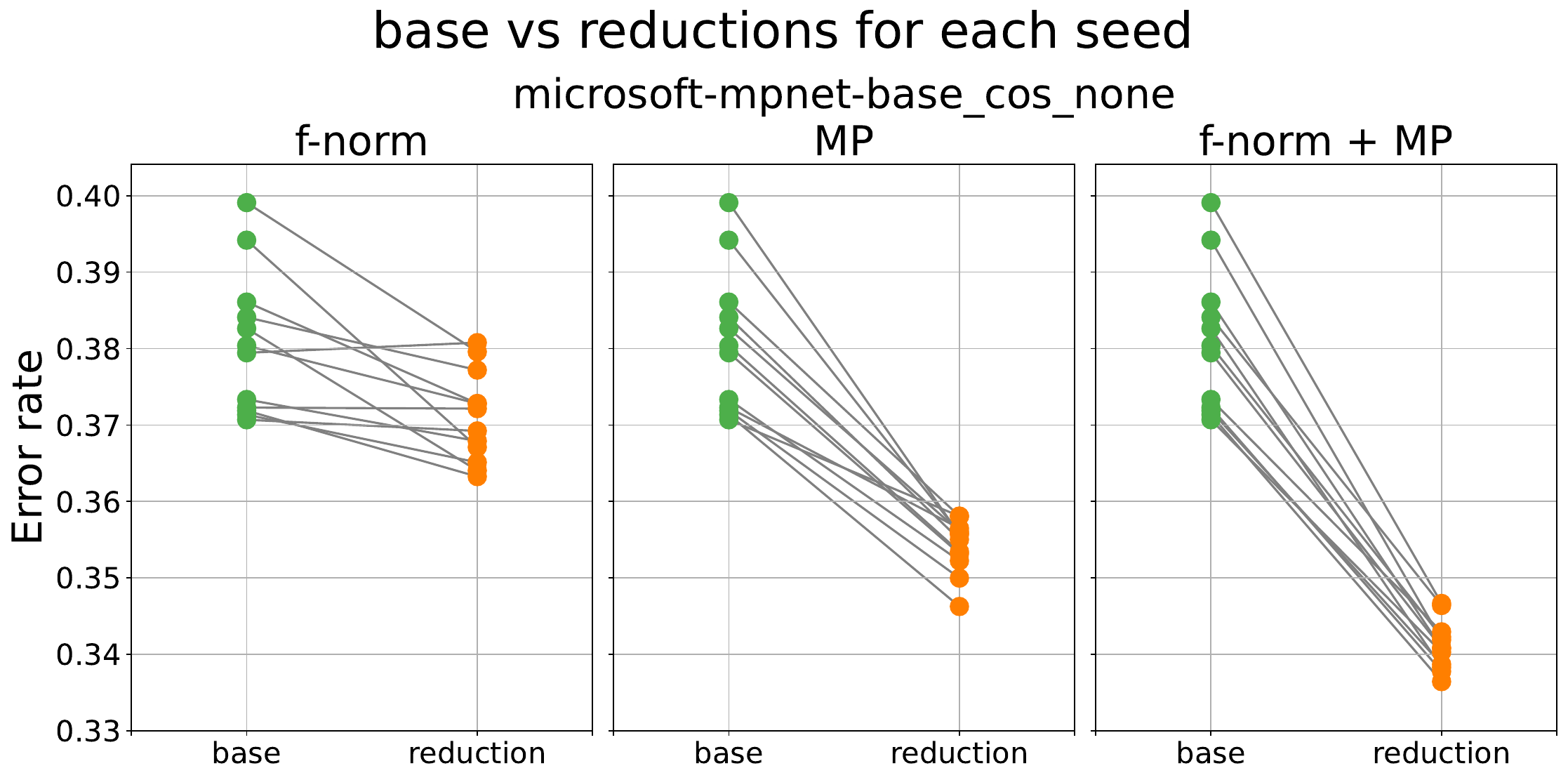}
\caption{20 Newsgroups. Average base k-skewness 4.79.}\label{fig:base_vs_reductions_microsoft-mpnet-base_cos}
\end{figure}

\begin{figure}[ht]
  \includegraphics[width=0.99\linewidth]{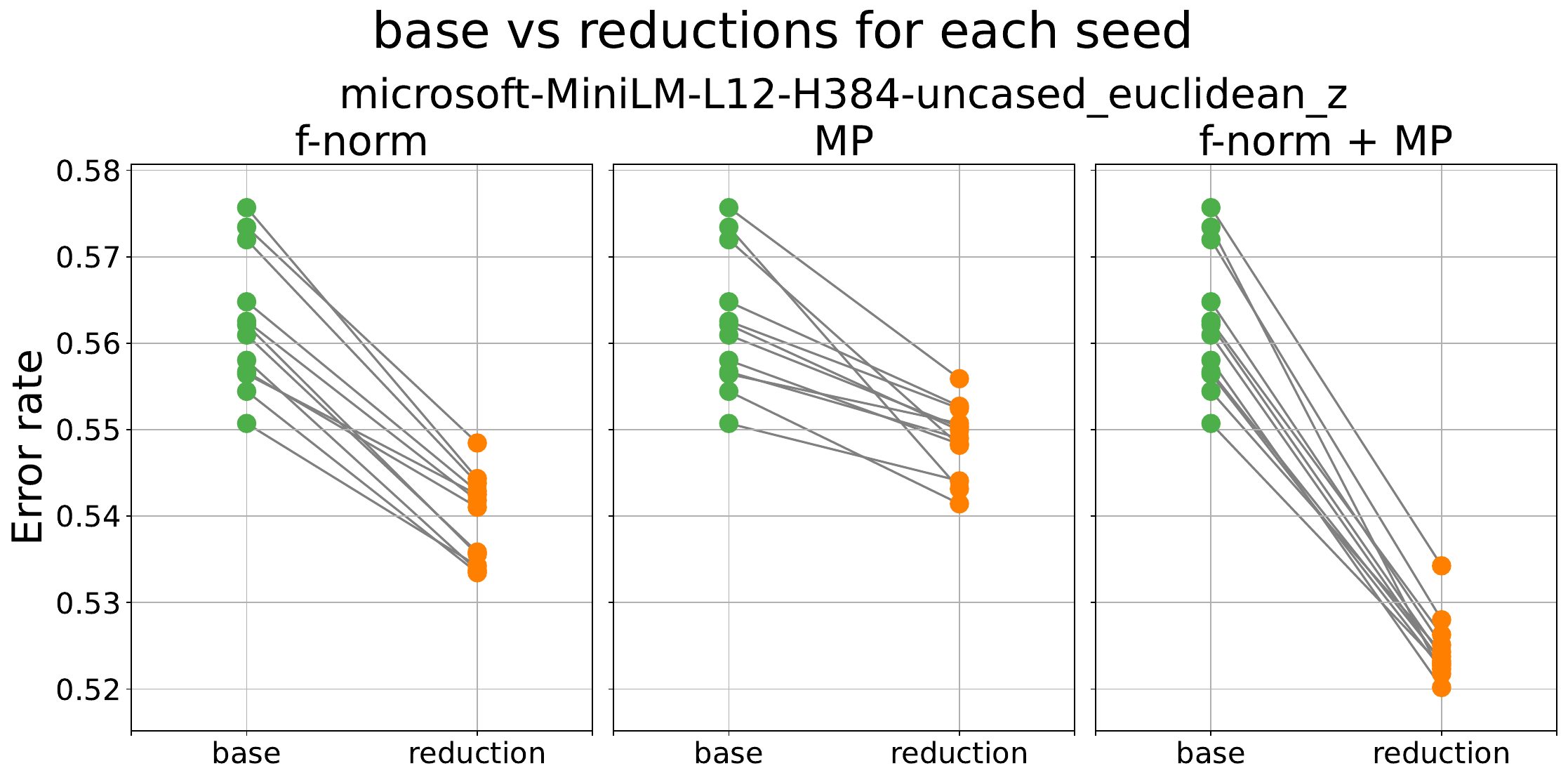}
\caption{20 Newsgroups. Average base k-skewness 10.06.}\label{fig:base_vs_reductions_microsoft-MiniLM-L12-H384-uncased_euc_z}
\end{figure}

\begin{figure}[ht]
  \includegraphics[width=0.99\linewidth]{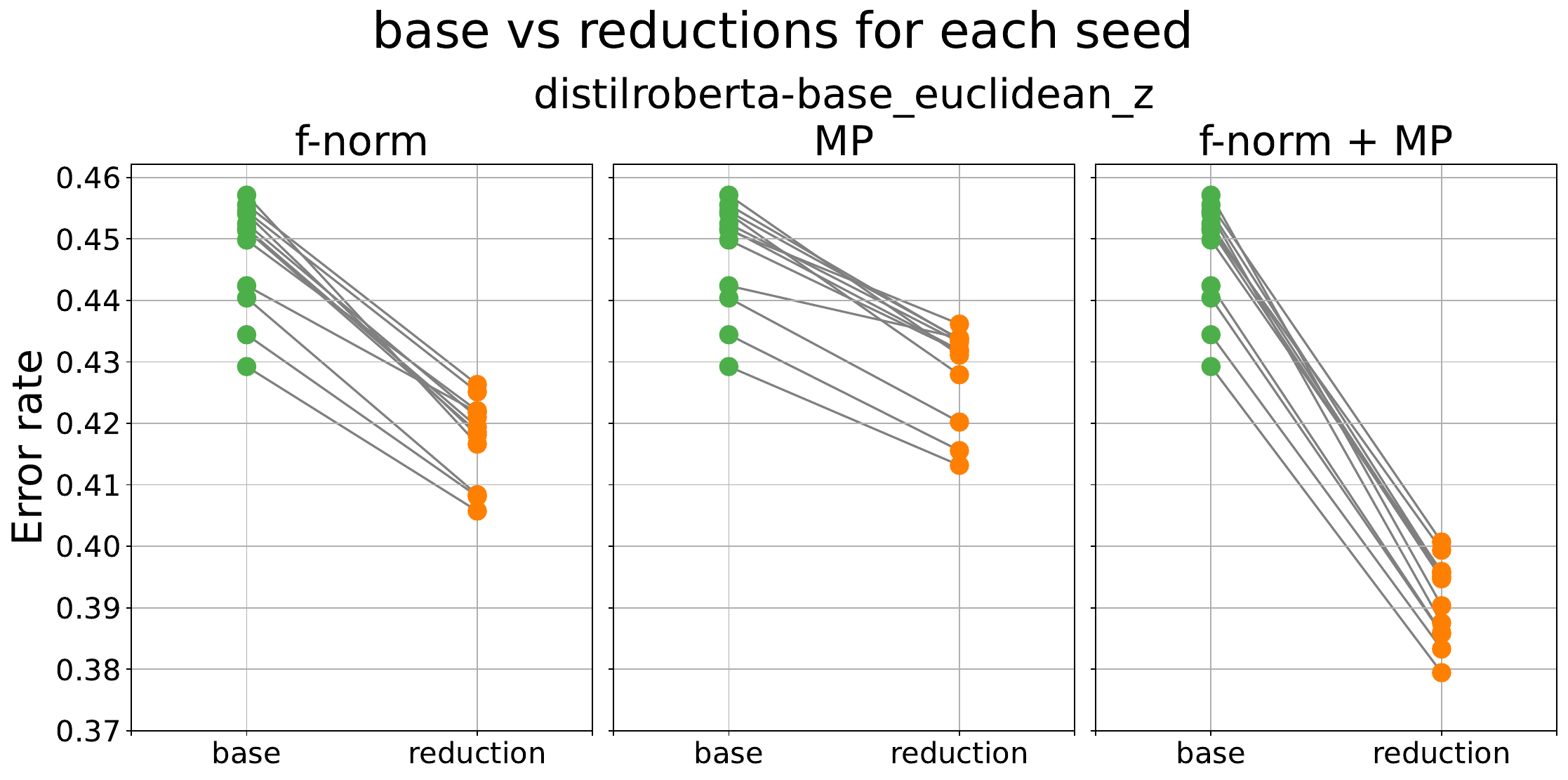}
\caption{20 Newsgroups. Average base k-skewness 5.92.}\label{fig:base_vs_reductions_distilroberta-base_euc_z}
\end{figure}

\begin{figure}[ht]
  \includegraphics[width=0.99\linewidth]{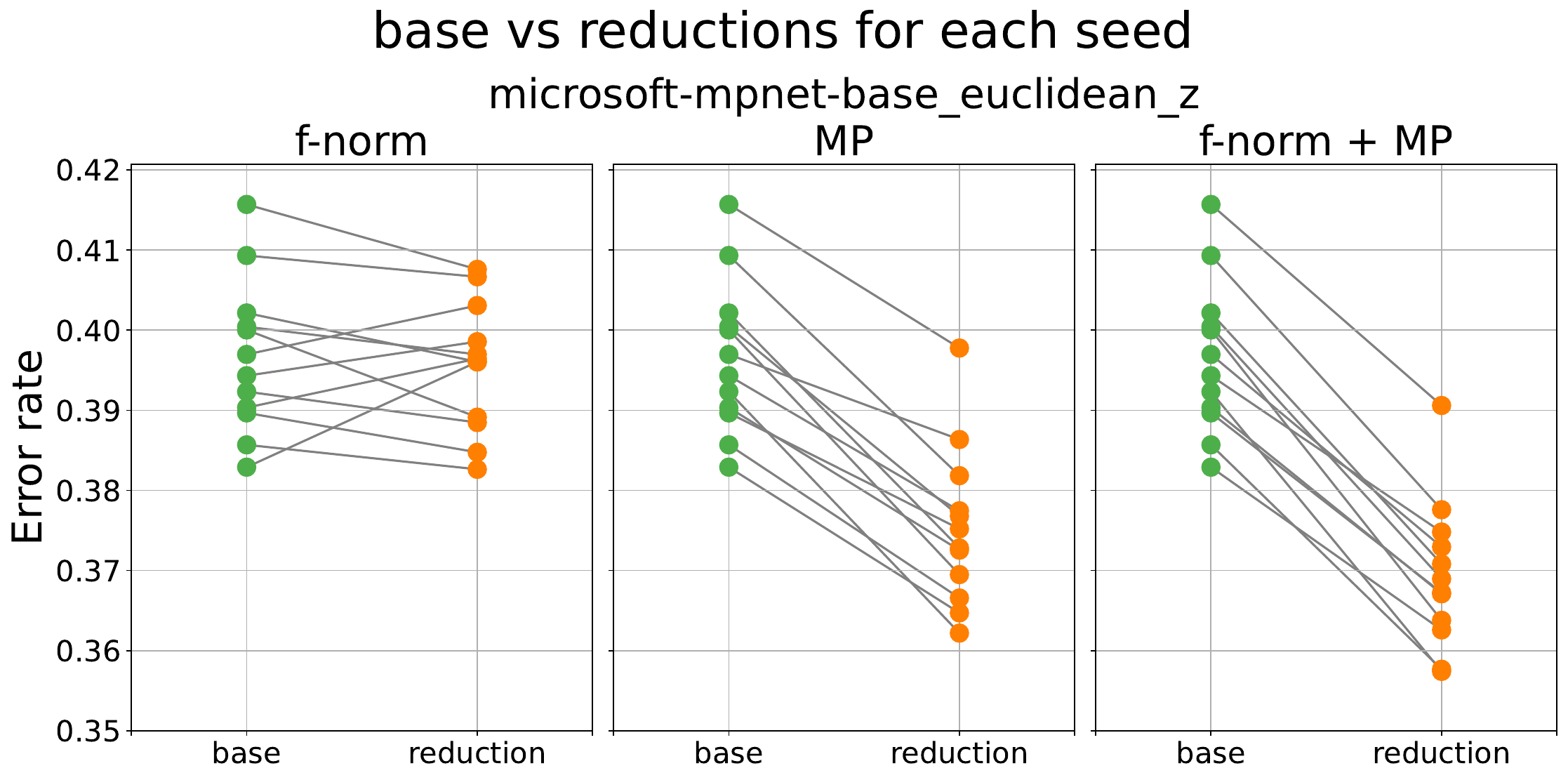}
\caption{20 Newsgroups. Average base k-skewness 5.50.}\label{fig:base_vs_reductions_microsoft-mpnet-base_euc_z}
\end{figure}

\begin{figure}[ht]
  \includegraphics[width=0.99\linewidth]{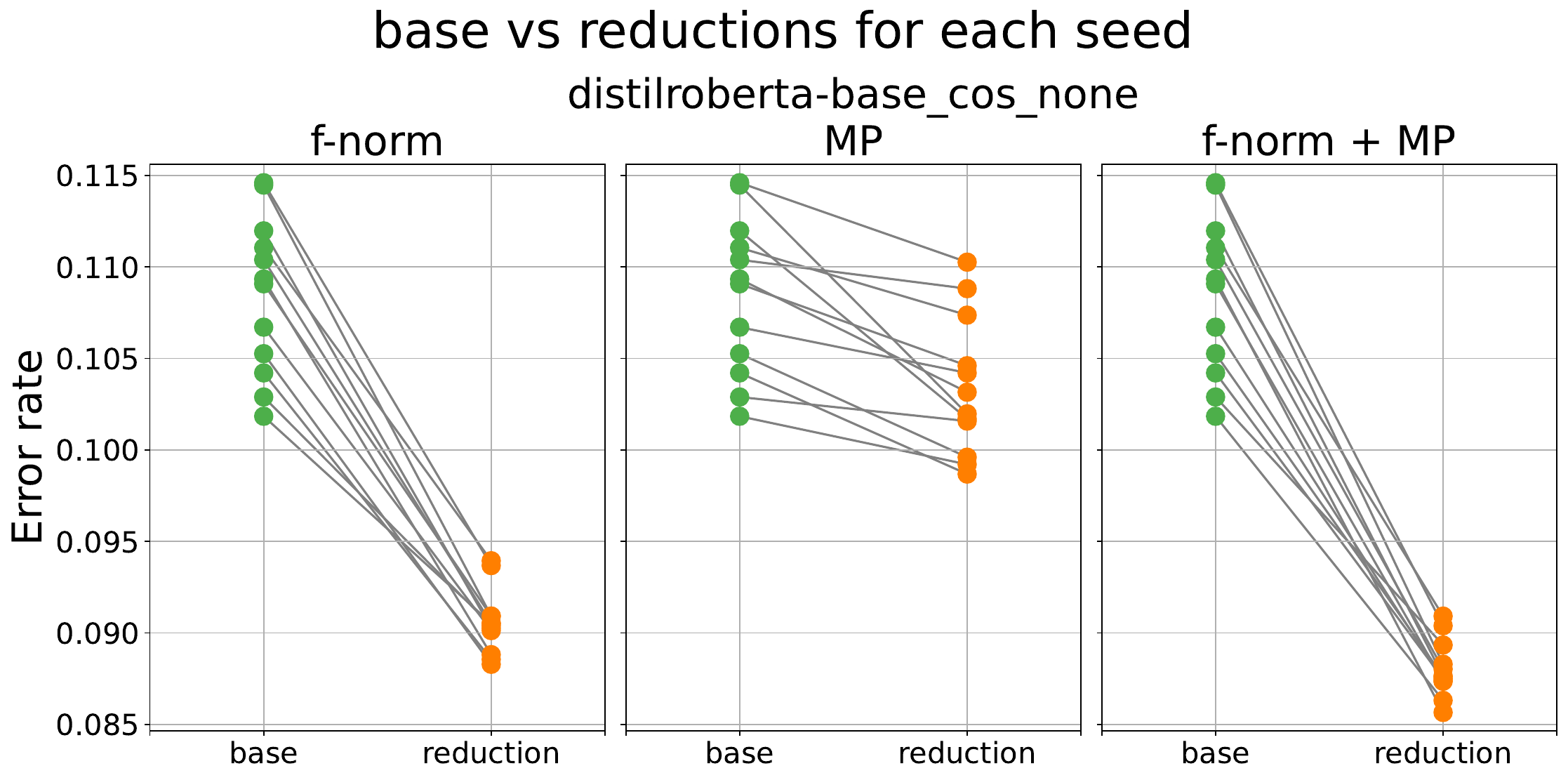}
\caption{AG News. Average base k-skewness 3.64.}\label{fig:base_vs_reductions_distilroberta-base_cos_ag}
\end{figure}

\begin{figure}[ht]
  \includegraphics[width=0.99\linewidth]{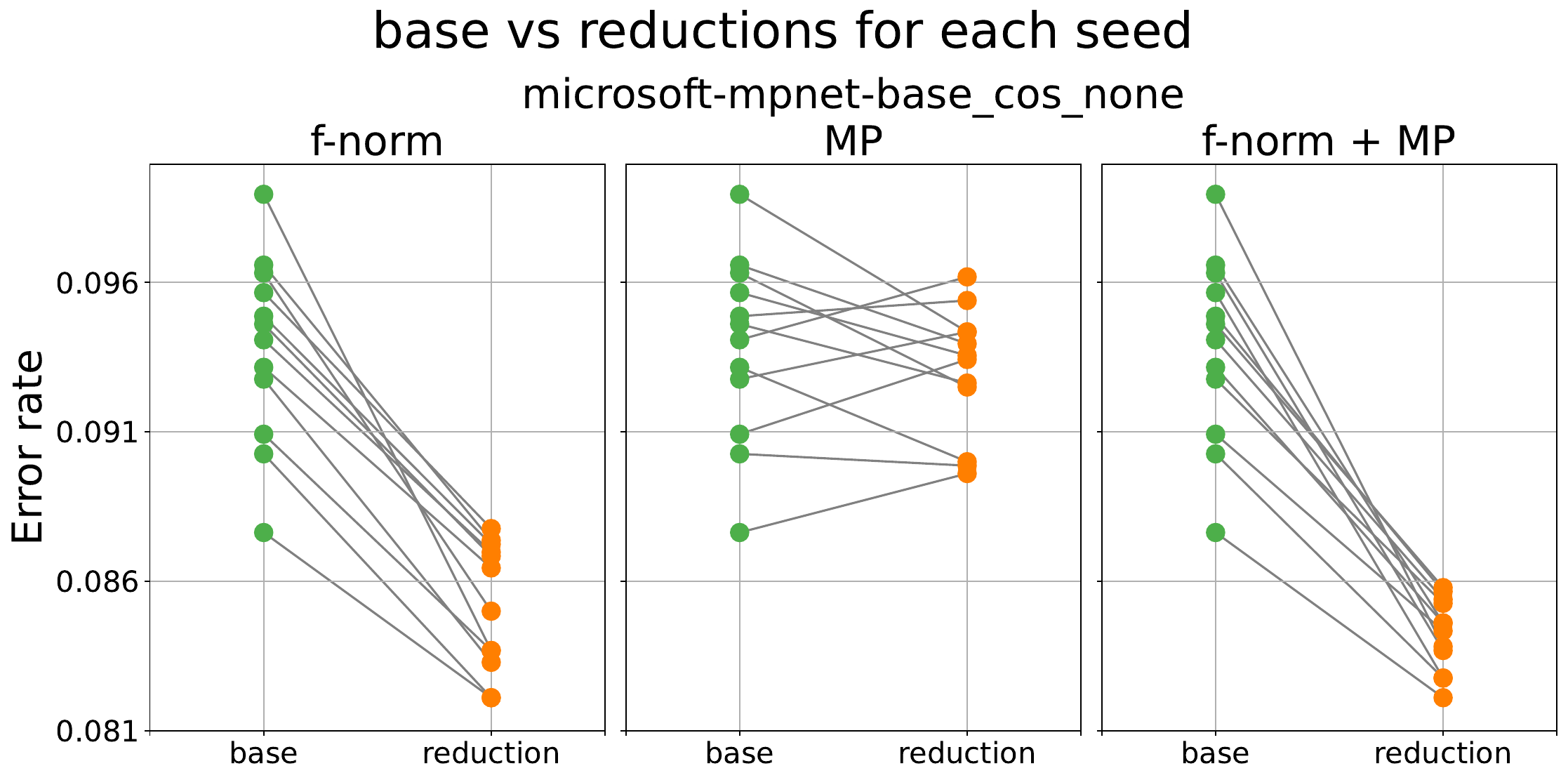}
\caption{AG News. Average base k-skewness 4.98.}\label{fig:base_vs_reductions_microsoft-mpnet-base_cos_ag}
\end{figure}

\begin{figure}[ht]
  \includegraphics[width=0.99\linewidth]{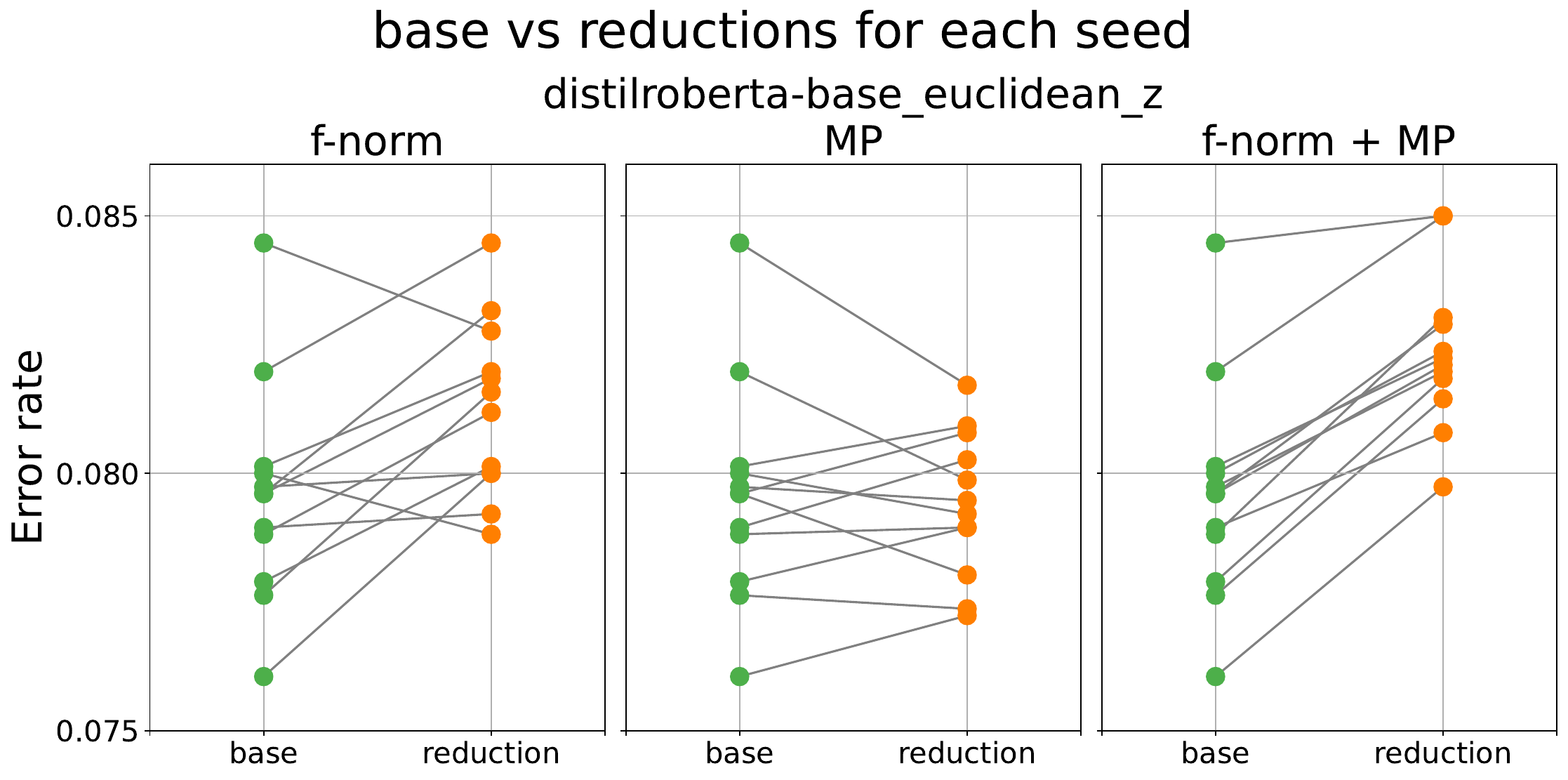}
\caption{AG News. Average base k-skewness 2.40.}\label{fig:base_vs_reductions_distilroberta-base_euc_z_ag}
\end{figure}

\begin{figure}[ht]
  \includegraphics[width=0.99\linewidth]{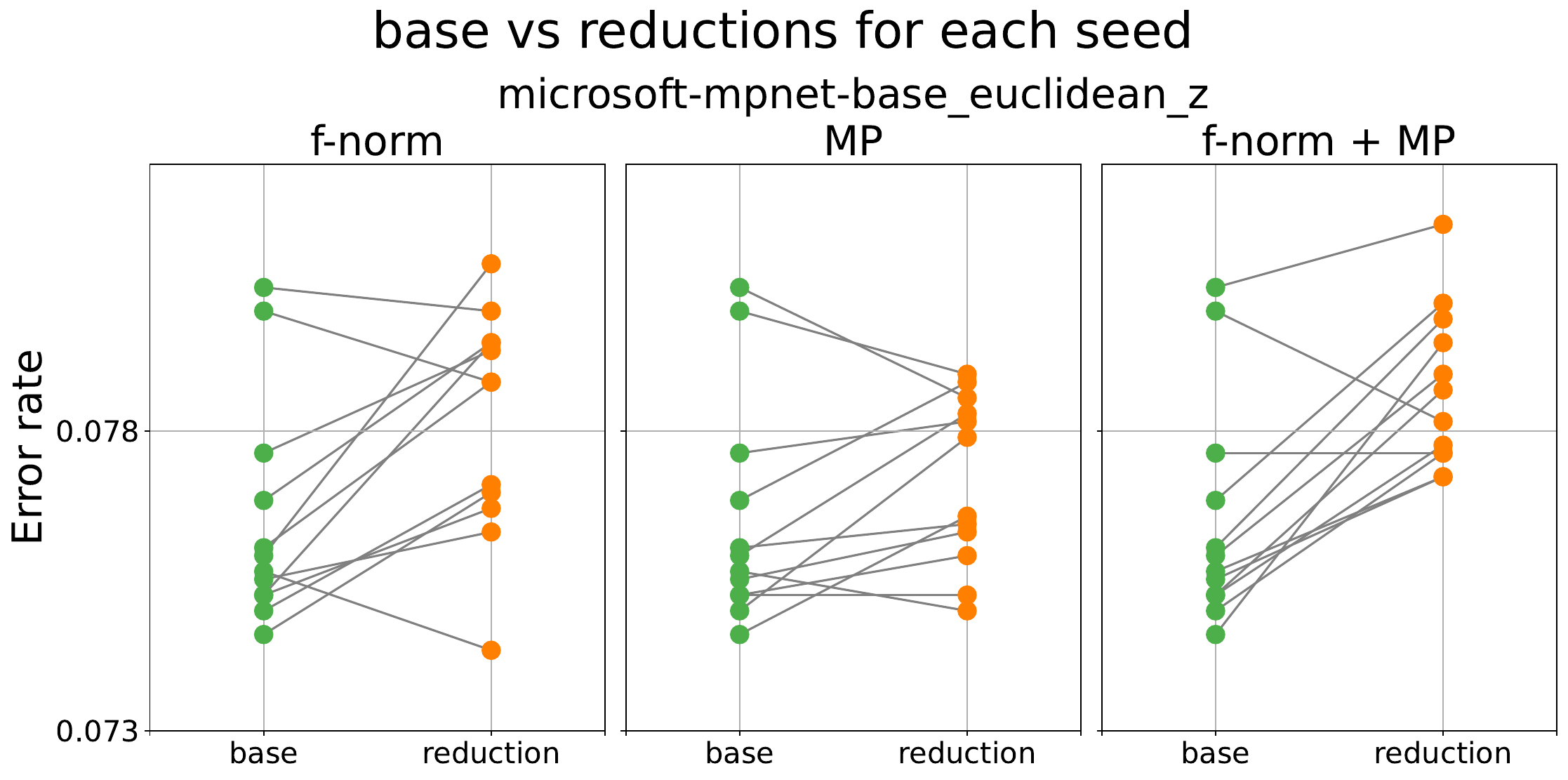}
\caption{AG News. Average base k-skewness 2.22.}\label{fig:base_vs_reductions_microsoft-mpnet-base_euc_z_ag}
\end{figure}

\begin{figure}[ht]
  \includegraphics[width=0.99\linewidth]{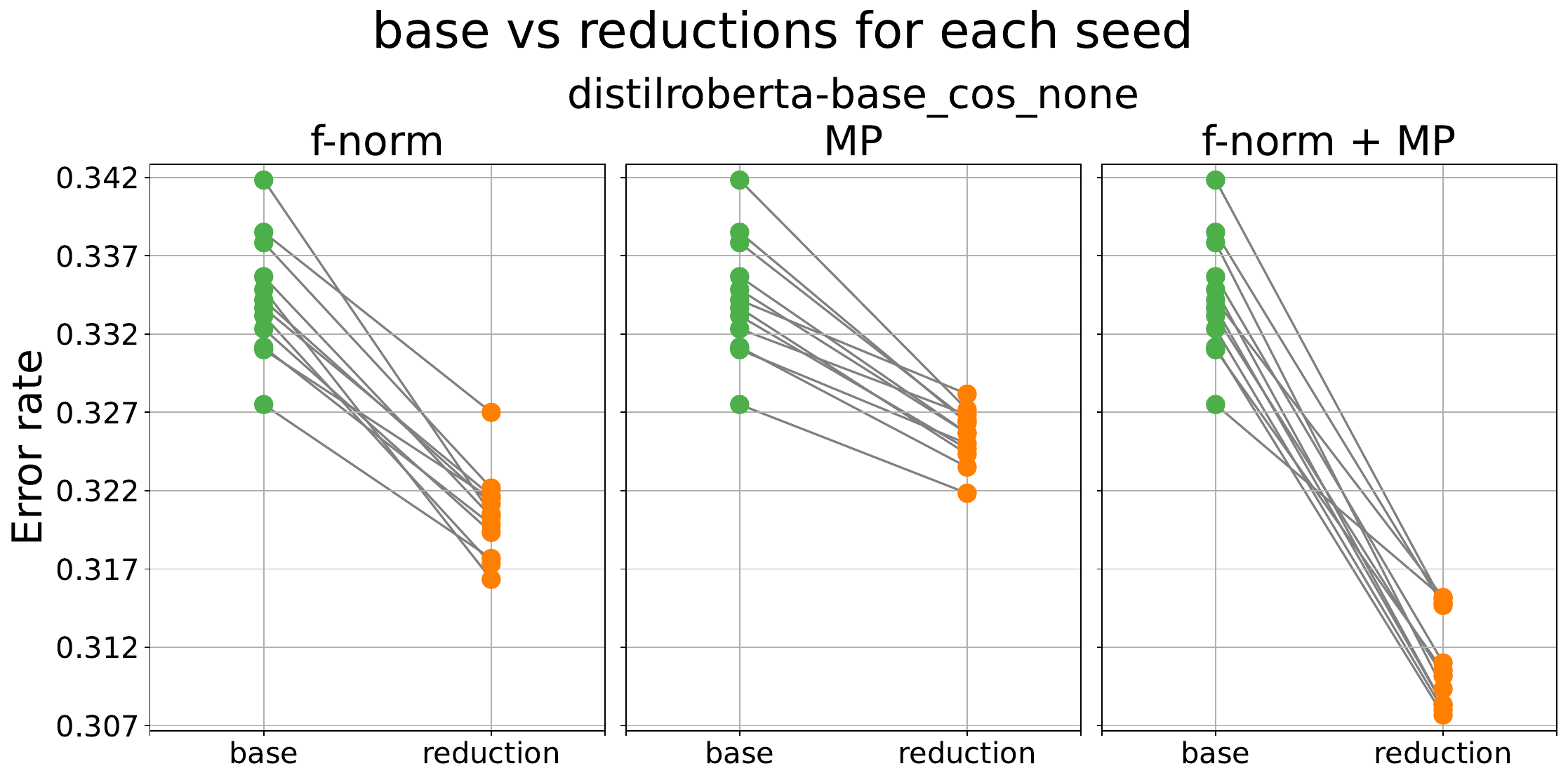}
\caption{Ten percent of Yahoo Answers. Average base k-skewness 8.41.}\label{fig:base_vs_reductions_distilroberta-base_cos_ya}
\end{figure}

\begin{figure}[ht]
  \includegraphics[width=0.99\linewidth]{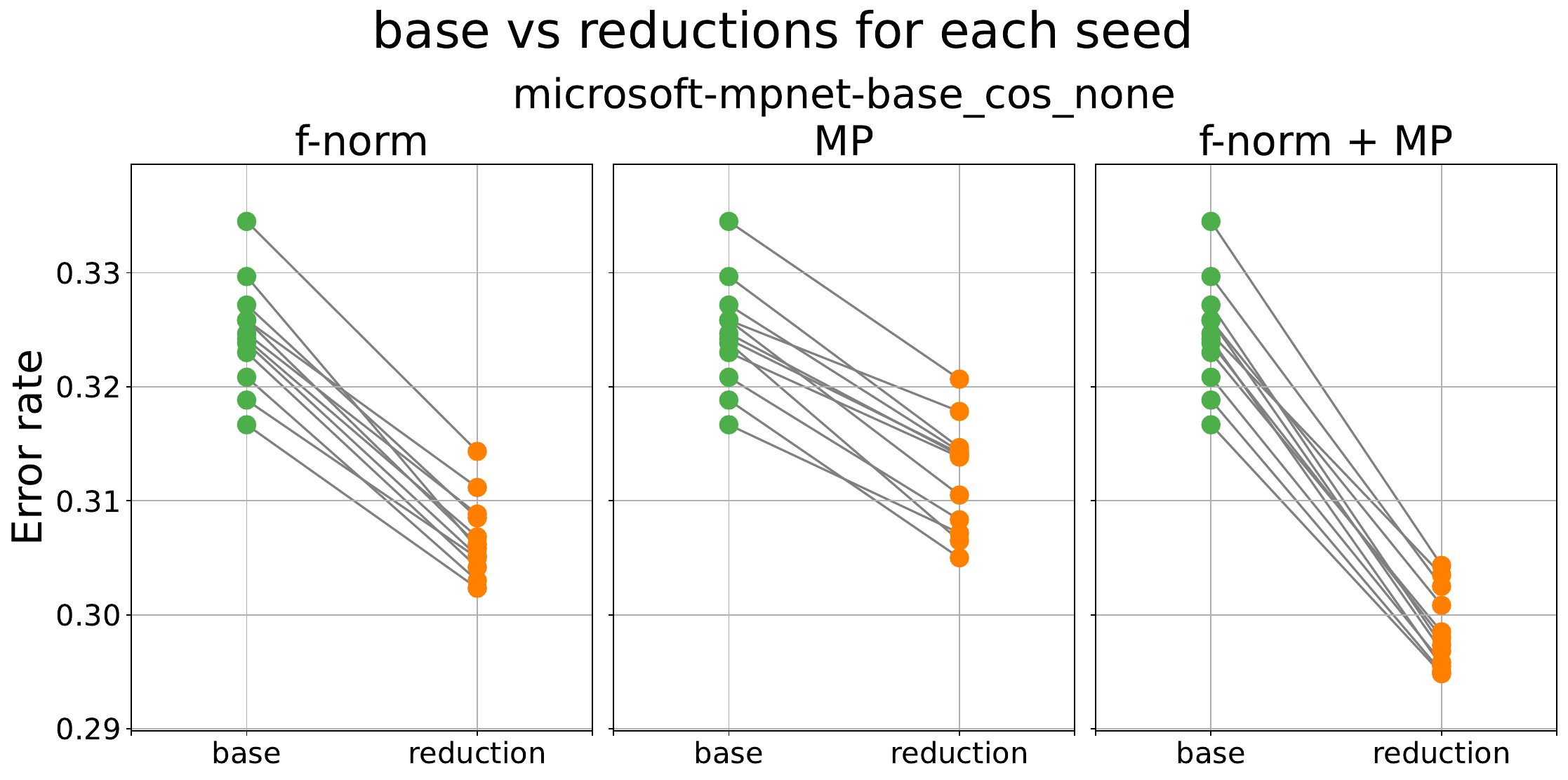}
\caption{Ten percent of Yahoo Answers. Average base k-skewness 5.54.}\label{fig:base_vs_reductions_microsoft-mpnet-base_cos_ya}
\end{figure}

\begin{figure}[ht]
  \includegraphics[width=0.99\linewidth]{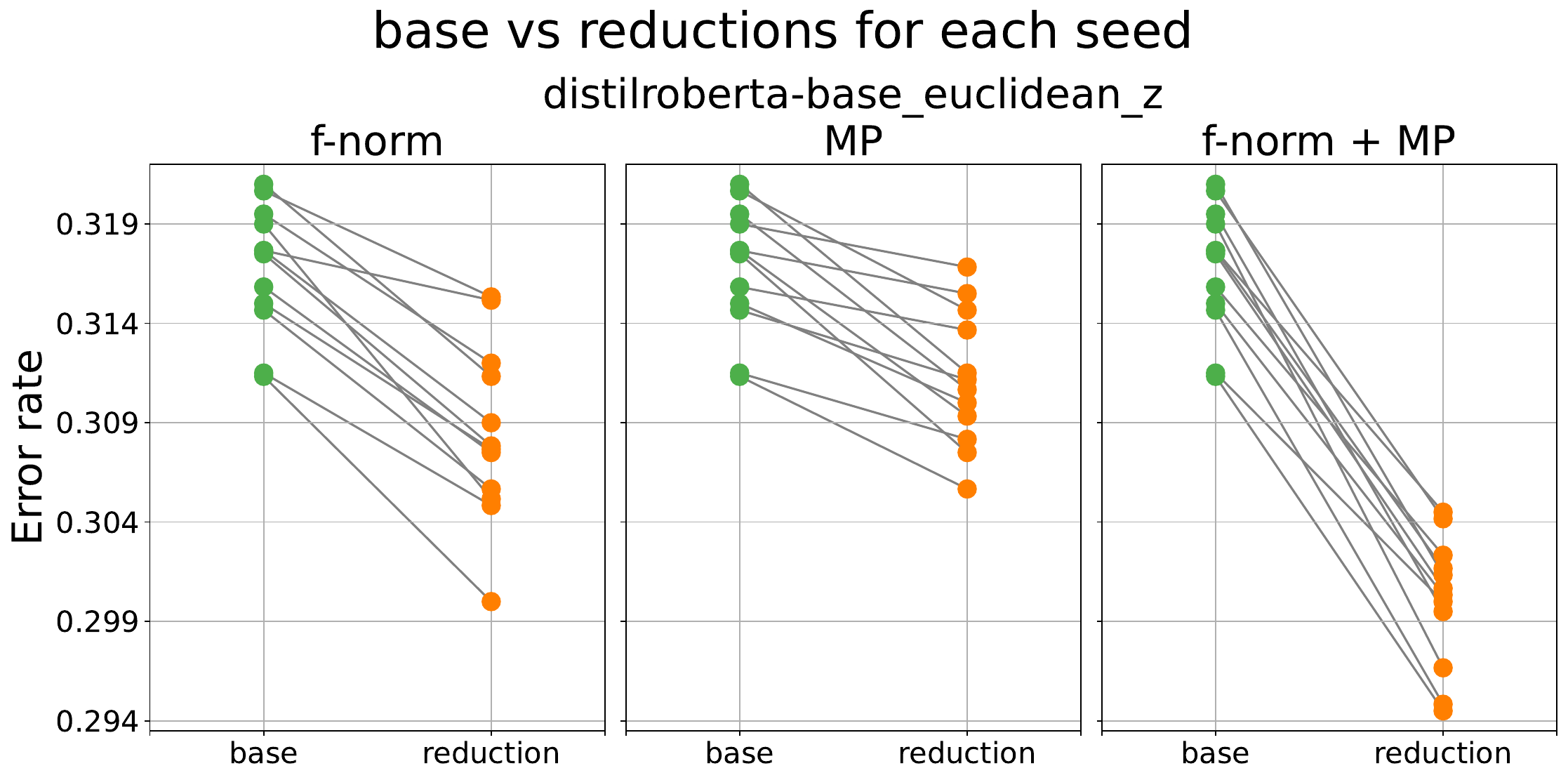}
\caption{Ten percent of Yahoo Answers. Average base k-skewness 3.29.}\label{fig:base_vs_reductions_distilroberta-base_euc_z_ya}
\end{figure}

\begin{figure}[ht]
  \includegraphics[width=0.99\linewidth]{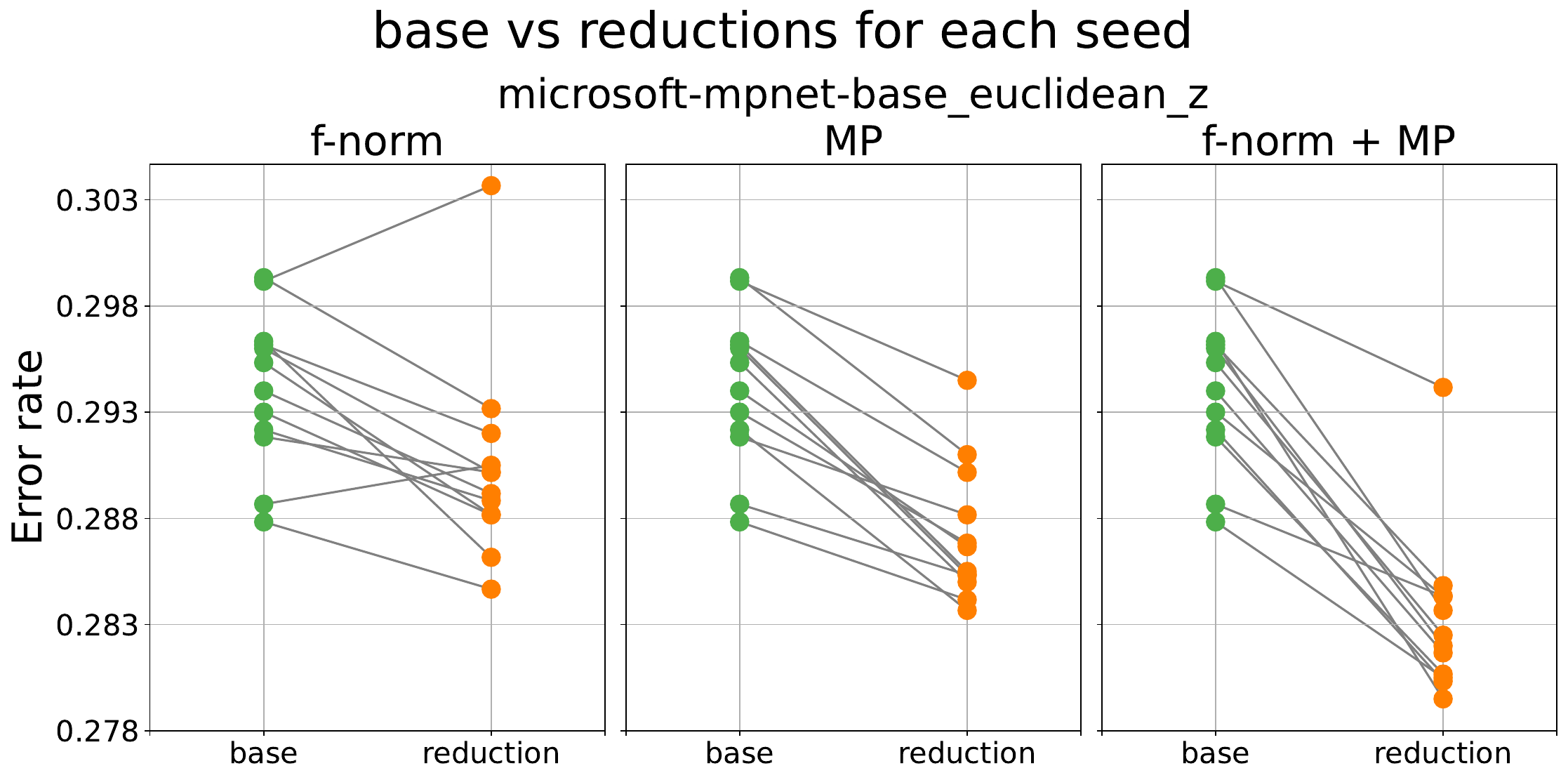}
\caption{Ten percent of Yahoo Answers. Average base k-skewness 3.05.}\label{fig:base_vs_reductions_microsoft-mpnet-base_euc_z_ya}
\end{figure}

Figures \ref{fig:base_vs_reductions_microsoft-MiniLM-L12-H384-uncased_cos}, \ref{fig:base_vs_reductions_distilroberta-base_cos}, \ref{fig:base_vs_reductions_microsoft-mpnet-base_cos}, \ref{fig:base_vs_reductions_microsoft-MiniLM-L12-H384-uncased_euc_z}, \ref{fig:base_vs_reductions_distilroberta-base_euc_z}, \ref{fig:base_vs_reductions_microsoft-mpnet-base_euc_z}, \ref{fig:base_vs_reductions_distilroberta-base_cos_ag}, \ref{fig:base_vs_reductions_microsoft-mpnet-base_cos_ag}, \ref{fig:base_vs_reductions_distilroberta-base_euc_z_ag}, \ref{fig:base_vs_reductions_microsoft-mpnet-base_euc_z_ag}, \ref{fig:base_vs_reductions_distilroberta-base_cos_ya}, \ref{fig:base_vs_reductions_microsoft-mpnet-base_cos_ya}, \ref{fig:base_vs_reductions_distilroberta-base_euc_z_ya}, \ref{fig:base_vs_reductions_microsoft-mpnet-base_euc_z_ya} show examples of how the error rate changes from the base case to cases where hubness reduction methods have been used.  






\section{Error rate - Without vs with Normalisation}
\label{sec:error_none_vs_n}
\begin{figure}[ht]
  \includegraphics[width=0.99\linewidth]{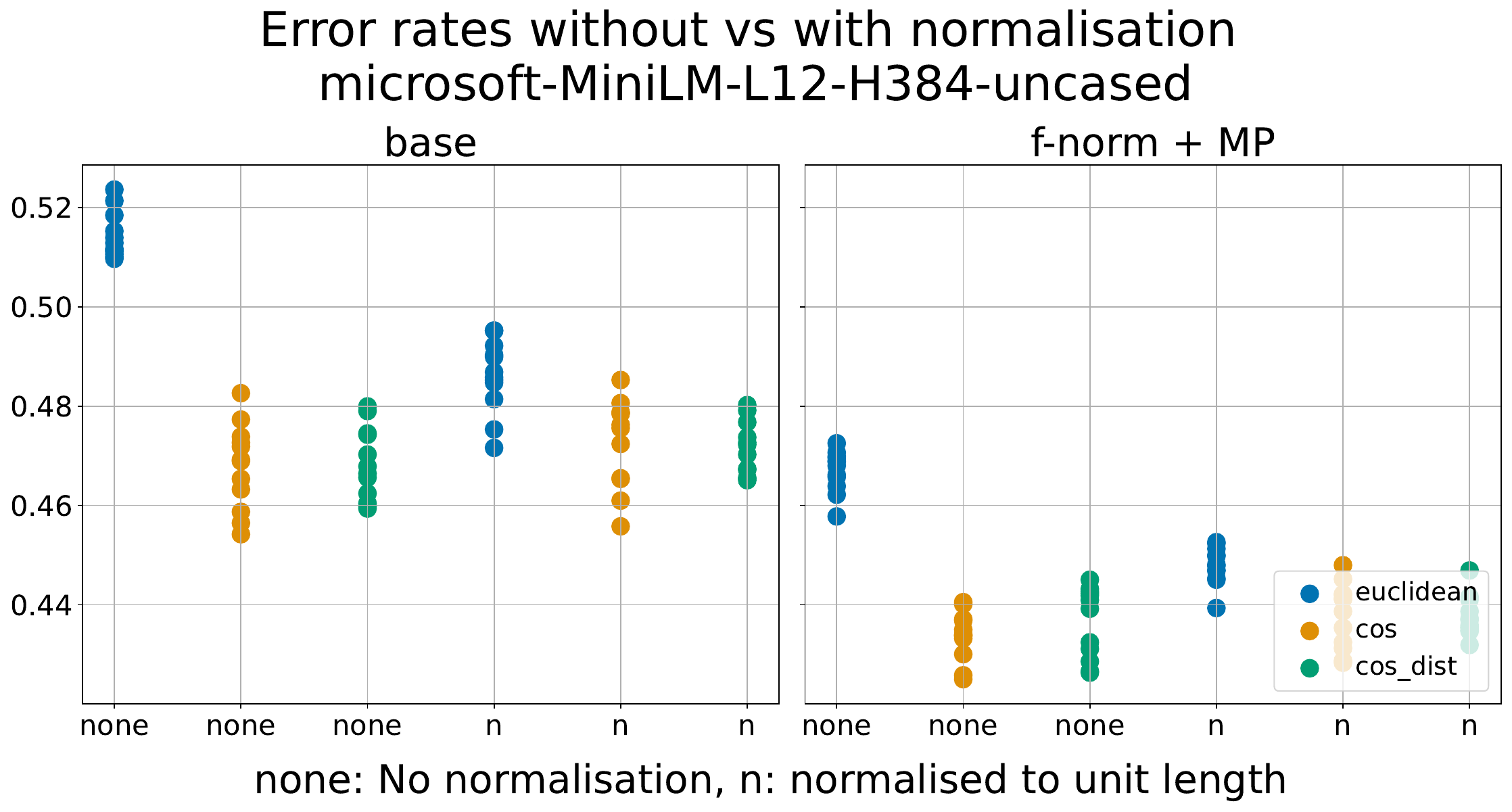}
\caption{20 Newsgroups.}\label{fig:none_vs_norm_microsoft-MiniLM-L12-H384-uncased}
\end{figure}

\begin{figure}[ht]
  \includegraphics[width=0.99\linewidth]{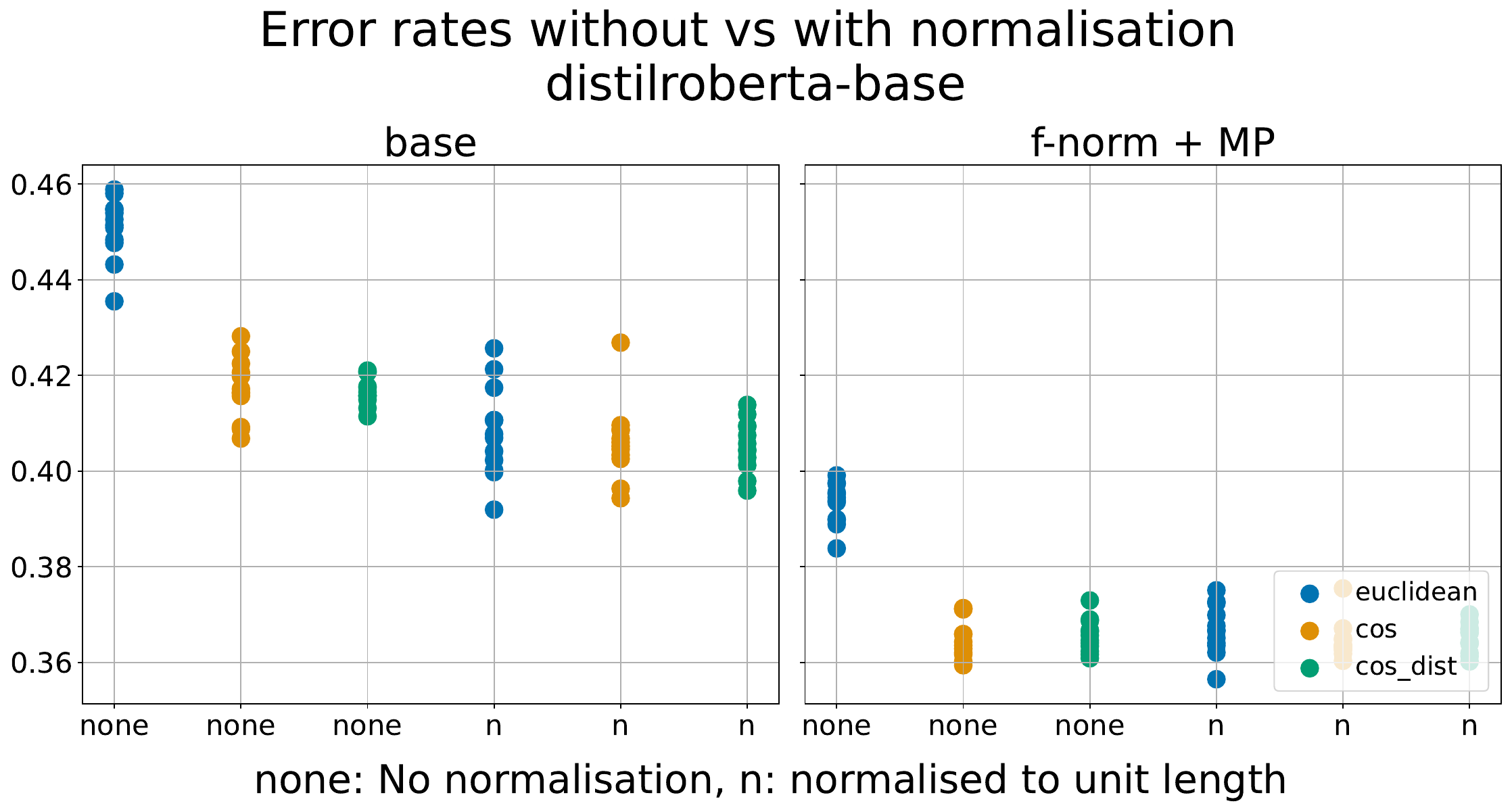}
\caption{20 Newsgroups.}\label{fig:none_vs_norm_distilroberta-base}
\end{figure}

\begin{figure}[ht]
  \includegraphics[width=0.99\linewidth]{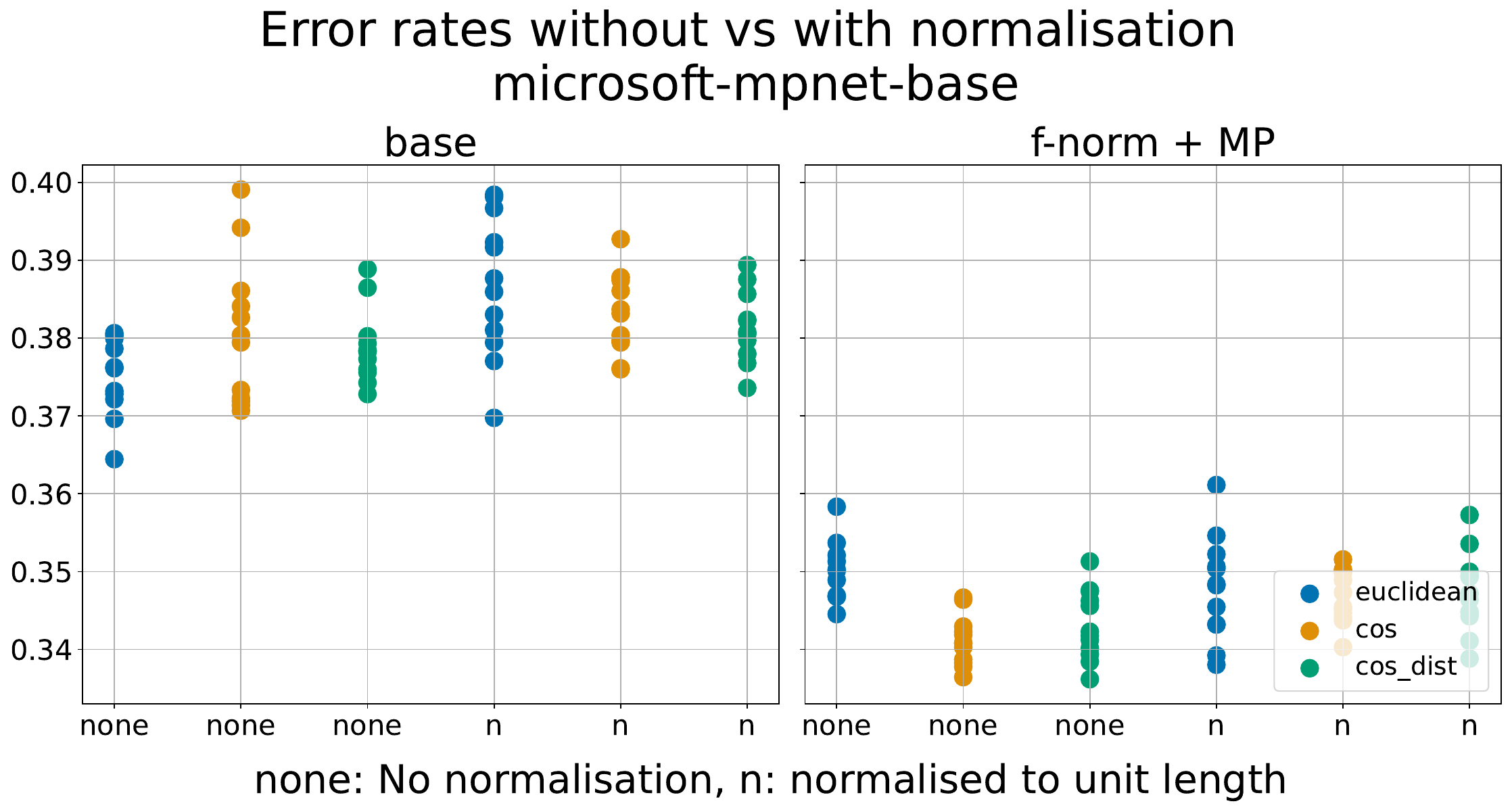}
\caption{20 Newsgroups. One model (euclidean) with greatly deviating values (base error rate about 80\%) has been removed to better show the variance in the remaining models.}\label{fig:none_vs_norm_microsoft-mpnet-base}
\end{figure}

\begin{figure}[ht]
  \includegraphics[width=0.99\linewidth]{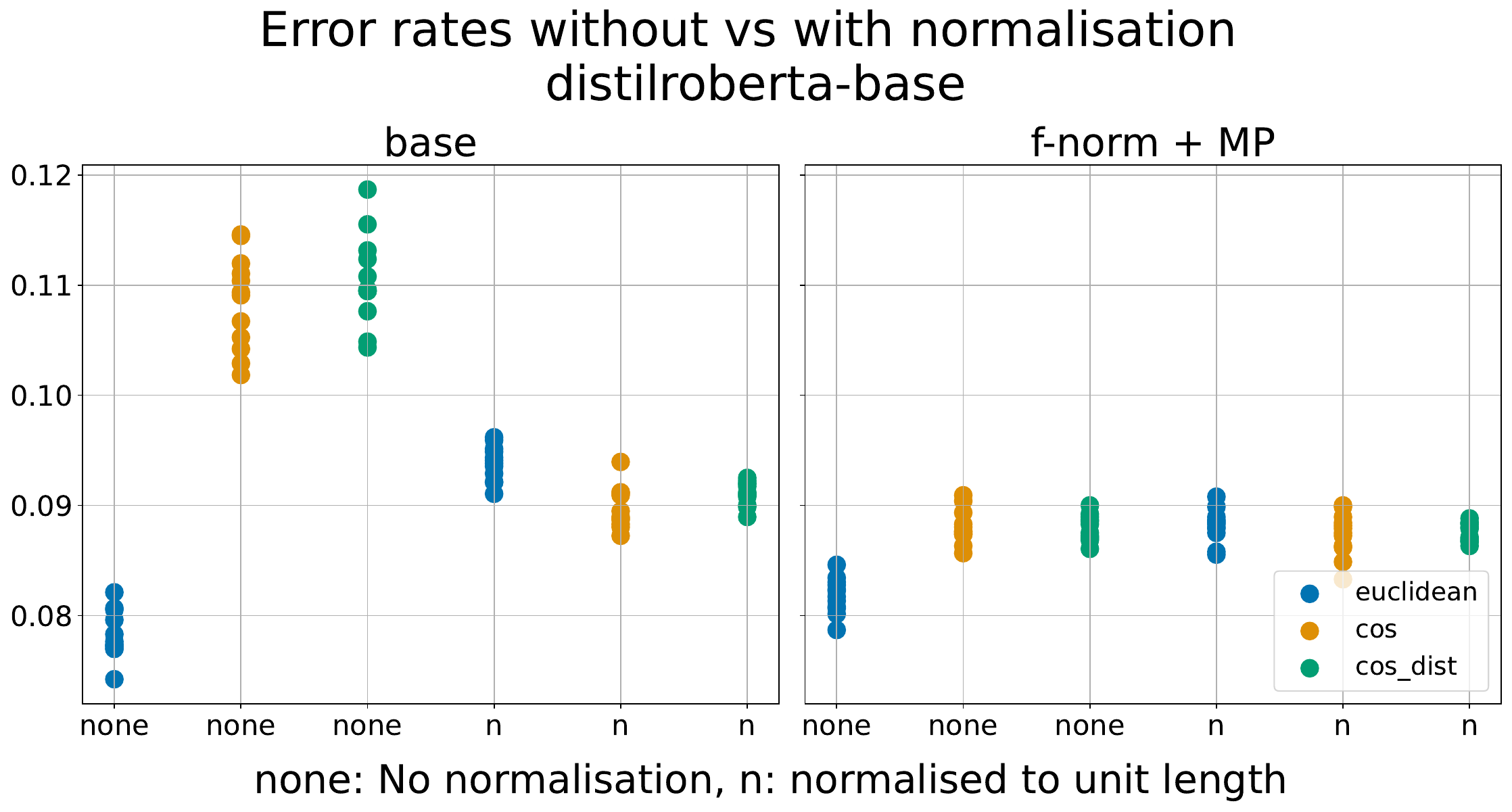}
\caption{AG News.}\label{fig:none_vs_norm_distilroberta-base_ag}
\end{figure}

\begin{figure}[ht]
  \includegraphics[width=0.99\linewidth]{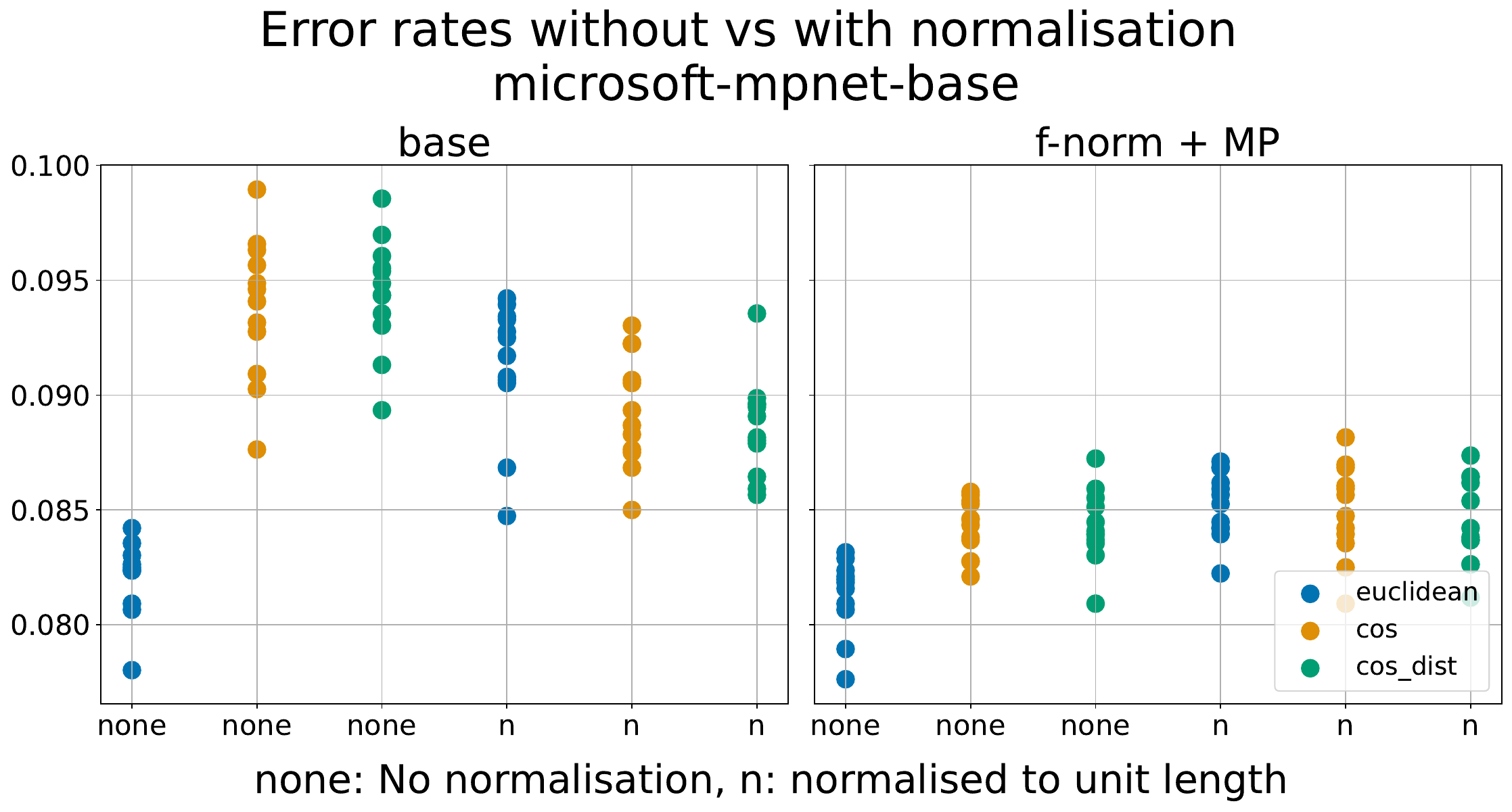}
\caption{AG News. One model (euclidean) with greatly deviating values (base error rate about 14\%) has been removed to better show the variance in the remaining models.}\label{fig:none_vs_norm_microsoft-mpnet-base_ag}
\end{figure}

\begin{figure}[ht]
  \includegraphics[width=0.99\linewidth]{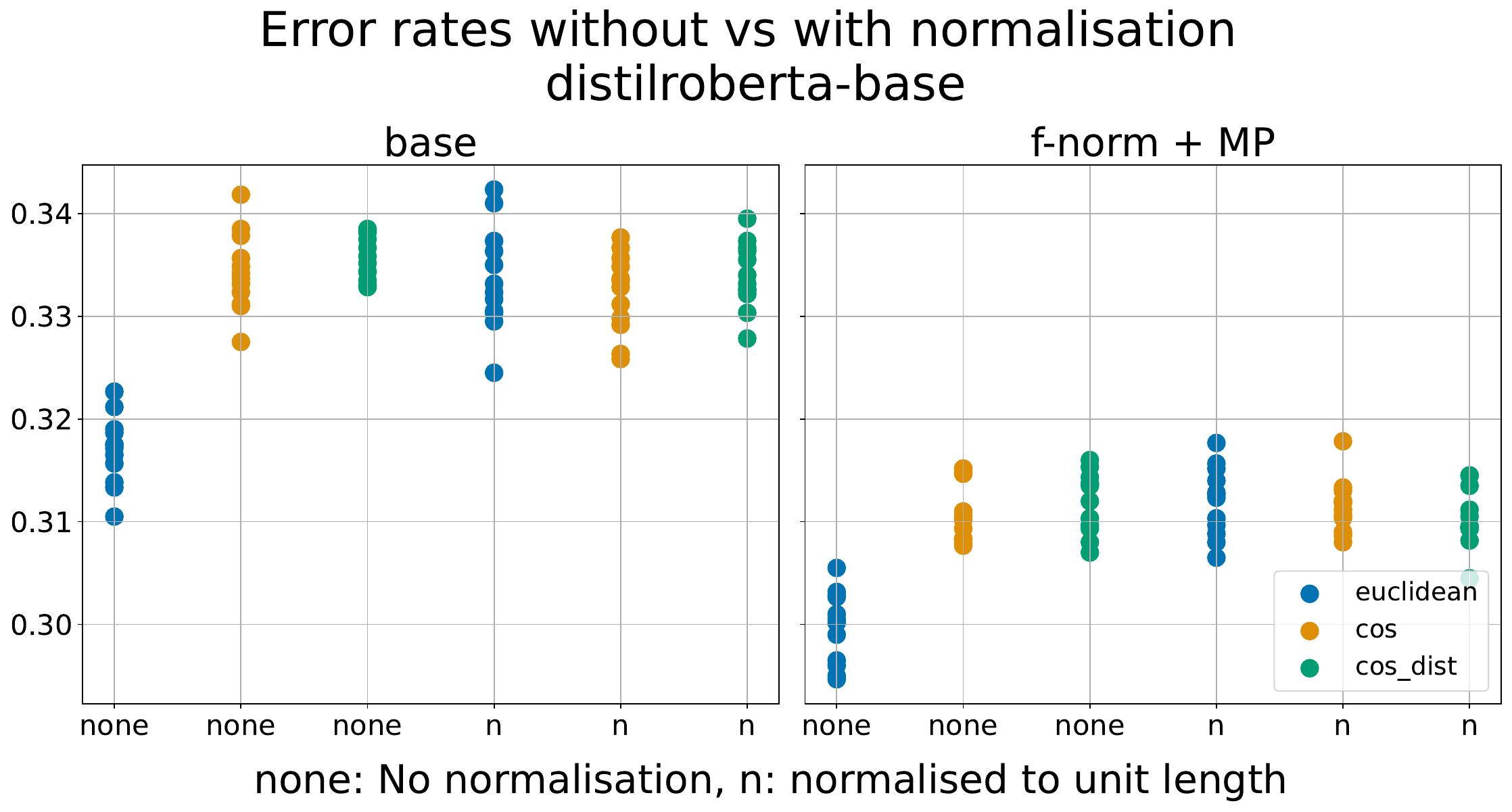}
\caption{Ten percent of Yahoo Answers.}\label{fig:none_vs_norm_distilroberta-base_ya}
\end{figure}

\begin{figure}[ht]
  \includegraphics[width=0.99\linewidth]{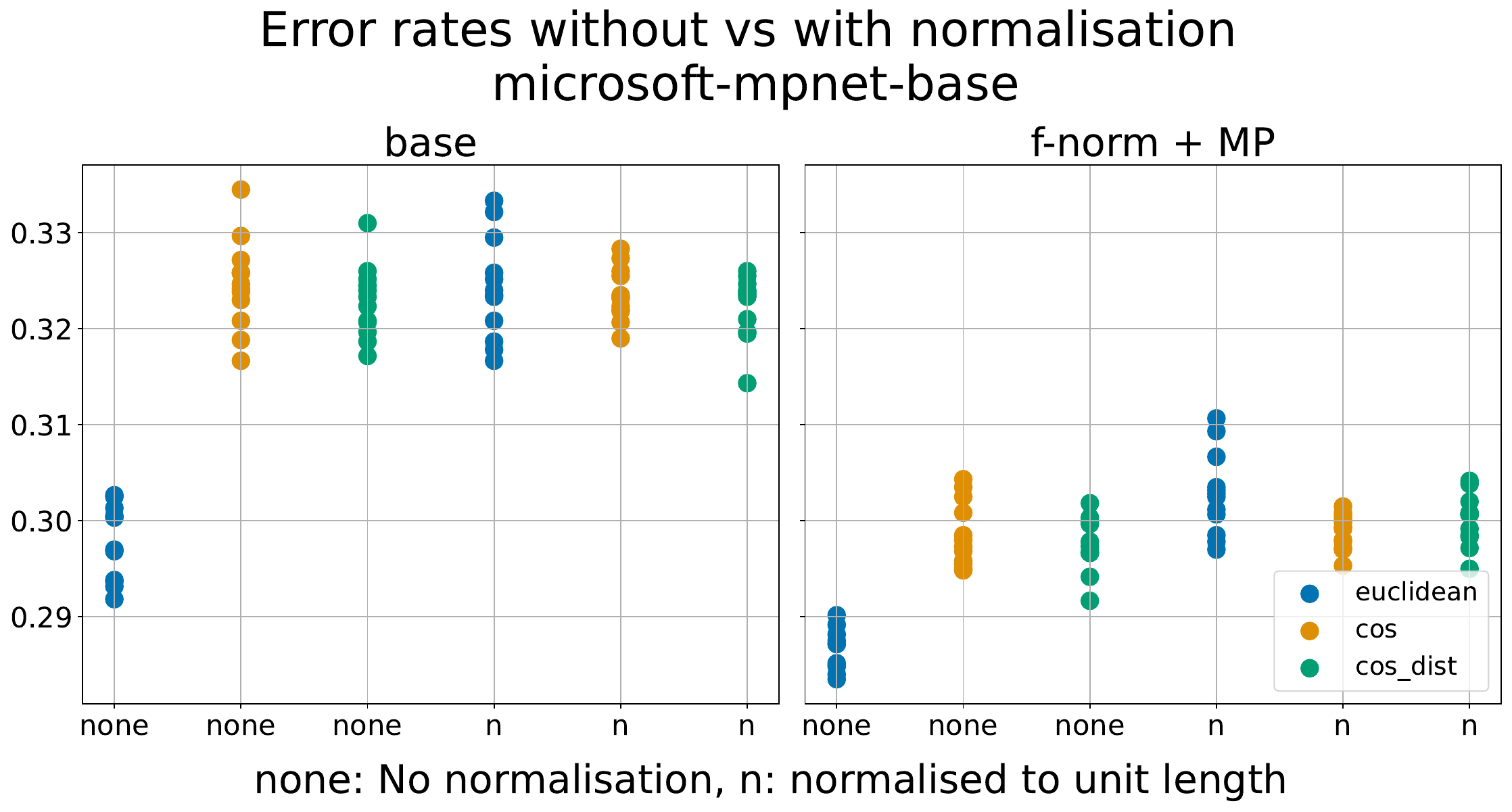}
\caption{Ten percent of Yahoo Answers. One model (euclidean) with greatly deviating values (base error rate about 55\%) has been removed to better show the variance in the remaining models.}\label{fig:none_vs_norm_microsoft-mpnet-base_ya}
\end{figure}

Figures \ref{fig:none_vs_norm_microsoft-MiniLM-L12-H384-uncased}, \ref{fig:none_vs_norm_distilroberta-base}, \ref{fig:none_vs_norm_microsoft-mpnet-base}, \ref{fig:none_vs_norm_distilroberta-base_ag}, \ref{fig:none_vs_norm_microsoft-mpnet-base_ag}, \ref{fig:none_vs_norm_distilroberta-base_ya}, \ref{fig:none_vs_norm_microsoft-mpnet-base_ya} show error rate differences for models which have been trained without and with normalisation to unit length before and after using f-norm + MP.




\section{Pretrained Models Results and Intended Use}
\label{sec:pretrained_models}
All pretrained models were made during "Community week using JAX/Flax for NLP \& CV"\footnote{\url{discuss.huggingface.co/t/train-the-best-sentence-embedding-model-ever-with-1b-training-pairs}}, organized by Hugging Face. All pretrained models used have Euclidean distance as a suitable score function, and they are all supposed to give semantic embeddings of sentences or short paragraphs. Table \ref{error_rate_hub_pretrained} shows effects on error rate and hubness when using f-norm, MP and f-norm + MP.

\begin{table*}
  \caption{\label{error_rate_hub_pretrained} Error rate (error), k-skewness (skew) and robinhood score (rh) on 20 Newsgroups (20), AG News (AG) and ten percent of Yahoo Answers (YA) for four pretrained models, base case vs after hubness reduction. Note that all these models have been trained on YA in various ways and their hubness on YA is in general lower than the models we trained. Thus, it seems that training on a specific dataset can make hubness go down for that dataset. However, hubness is still high on 20 which the models have not been trained on.}
  \centering
  \begin{tabular}{l | c c c | c c c | c c c}
    \hline
    & \multicolumn{3}{c |}{\textbf{20}} & \multicolumn{3}{c |}{\textbf{AG}} & \multicolumn{3}{c}{\textbf{YA}}          \\
    \hline
    & \textbf{error} & \textbf{skew} & \textbf{rh} & \textbf{error} &  \textbf{skew} & \textbf{rh} & \textbf{error} & \textbf{skew} & \textbf{rh} \\
    \hline
    \multicolumn{4}{l |}{\textbf{multi-qa-distilbert-cos-v1}} & & & & & & \\ 
    \hline
\textbf{base}         & 0.318 & 7.58  & 0.37 & 0.081 & 1.90 & 0.59 & 0.271 & 2.12 & 0.69 \\
 \textbf{f-norm}      & 0.316 & 2.15  & 0.34 & 0.082 & 1.88 & 0.58 & 0.268 & 2.16 & 0.69 \\
 \textbf{MP}          & 0.292 & 13.06 & 0.28 & 0.081 & 1.55 & 0.55 & 0.271 & 1.77 & 0.66 \\
 \textbf{f-norm + MP} & 0.289 & 1.89  & 0.30 & 0.079 & 1.57 & 0.56 & 0.269 & 1.88 & 0.67 \\
\hline
\multicolumn{4}{l |}{\textbf{all-MiniLM-L12-v2}} & & & & & & \\
\hline
\textbf{base}        & 0.324 & 11.45 & 0.35 & 0.088 & 1.91 & 0.58 & 0.268 & 1.98 & 0.68 \\
\textbf{f-norm}      & 0.324 & 2.22  & 0.33 & 0.086 & 1.88 & 0.58 & 0.269 & 1.97 & 0.68 \\
\textbf{MP}          & 0.302 & 16.24 & 0.28 & 0.086 & 1.55 & 0.55 & 0.270 & 1.78 & 0.66 \\
\textbf{f-norm + MP} & 0.300 & 1.88  & 0.29 & 0.085 & 1.56 & 0.56 & 0.267 & 1.79 & 0.67 \\
\hline
\multicolumn{4}{l |}{\textbf{all-mpnet-base-v2}} & & & & & & \\
\hline
 \textbf{base}       & 0.295 & 11.53 & 0.37 & 0.082 & 1.81 & 0.58 & 0.251 & 2.02 & 0.68 \\
  \textbf{f-norm}    & 0.293 &  2.32 & 0.34 & 0.080 & 1.80 & 0.58 & 0.256 & 1.96 & 0.68 \\
 \textbf{MP}         & 0.272 & 15.80 & 0.30 & 0.085 & 1.57 & 0.55 & 0.250 & 1.73 & 0.66 \\
\textbf{f-norm + MP} & 0.272 &  1.98 & 0.30 & 0.082 & 1.54 & 0.56 & 0.250 & 1.78 & 0.67 \\
\hline
\multicolumn{4}{l |}{\textbf{all-distilroberta-v1}} & & & & & & \\
\hline
\textbf{base}          & 0.322 &  6.61 & 0.39 & 0.085 & 1.93 & 0.59 & 0.254 & 2.08 & 0.68 \\
 \textbf{f-norm}       & 0.315 &  2.53 & 0.35 & 0.084 & 1.90 & 0.58 & 0.257 & 2.07 & 0.68 \\
  \textbf{MP}          & 0.291 & 11.20 & 0.29 & 0.087 & 1.57 & 0.55 & 0.254 & 1.72 & 0.66 \\
  \textbf{f-norm + MP} & 0.296 &  2.05 & 0.31 & 0.085 & 1.59 & 0.56 & 0.253 & 1.82 & 0.67 \\
    \hline
  \end{tabular}
\end{table*}

\section{Computational Experiment Details}
\label{sec:comp_details}

\subsection{Hyperparameters}
When training our models, we kept the default hyperparameters for the SentenceTransformer object as can be found at \url{github.com/UKPLab/sentence-transformers/blob/master/sentence_transformers/SentenceTransformer.py}. Thus, the optimizer was AdamW with a learning rate of $2\cdot10^{-5}$, weight decay was $0.01$, and the learning rate schedular was set to WarmupLinear. We did not do a hyperparameter search, since that was not the point of our experiment. 

\subsection{GPU and CPU Types}
Training of the models and making embeddings was done on the following GPU types: NVIDIA TITAN Xp, GeForce GTX 1080 Ti, NVIDIA TITAN V and NVIDIA TITAN X. \par 
The knn and hubness measures were done on CPU on either AMD Ryzen 7 PRO 5850U or Xeon E5-2620 v4.

\subsection{Computation Time}
Training and generating embeddings times reported for running on an NVIDIA TITAN Xp GPU. \par
The total training time of the 540 models was about 28.5 hours. Of this, the models using microsoft-mpnet-base as base model took about 12 hours, the models using distilroberta-base as base model took about 6 hours and the models using microsoft-MiniLM-L12-H384-uncased as base model took about 10.5 hours. Note that a single model takes 4 minutes or less to train. \par 
Total time generating embeddings for our own models was about 11.50 hours for 20 Newsgroups for all 540 models, about 21 hours for AG News for the 360 models using microsoft-mpnet-base or distilroberta-base as base model, and about 46.5 hours for the ten percent of Yahoo Answers for the 360 models using microsoft-mpnet-base or distilroberta-base as base model. \par 
Generating embeddings for the four pretrained models took about an hour. This gives us a total of 108.5 hours of GPU time for training and embeddings. \par 
Times for knn and hubness measures reported for running on an AMD Ryzen 7 PRO 5850U CPU. \par 
Getting knn results for our own models took about 69 hours for 20 Newsgroups for all 540 models, about 267 hours for AG News for the 360 models using microsoft-mpnet-base or distilroberta-base as base model, and about 306 hours for the ten percent of Yahoo Answers for the 360 models using microsoft-mpnet-base or distilroberta-base as base model. \par 
Getting knn results from the pretrained models for all datasets took about 7 hours. \par 
Getting the hubness results for our own models took about 7.5 hours for 20 Newsgroups for all 540 models, about 46.5 hours for AG News for the 360 models using microsoft-mpnet-base or distilroberta-base as base model, and about 60 hours for the ten percent of Yahoo Answers for the 360 models using microsoft-mpnet-base or distilroberta-base as base model. \par 
Getting hubness results from the pretrained models for all datasts took about 1.5 hours. \par 
Thus, in total we used 764.5 hours of CPU time for knn and hubness measures.

\section{Package Versions}
\label{sec:versions}
Important packages: h5py 3.7.0, hdf5 1.10.6, nltk 3.7, numpy 1.21.5, python 3.9.7, pytorch 1.12.0, sci-kit-hubness 0.30.0a1, scikit-learn 1.1.1, sentence-transformers 2.2.2, transformers 4.20.1. Complete environment description can be found on github\footnote{\url{https://github.com/bemigini/hubness-reduction-sentence-bert}}.

\section{Training on STS benchmark}
\label{sec:sts_details}
 The STS benchmark dataset has pairs of sentences annotated with a semantic similarity score from 0 (not similar) to 5 (semantically equal). It is split into a train, a dev, and a test set with $7251$, $1500$, and $2886$ sentence pairs, respectively. We trained our models on the train split. License: Unknown. \par
 When training the models, we had to decide what distances these scores should correspond to. There were two natural choices: The maximum distance between embeddings should be the distance between orthogonal unit vectors, $\sqrt{2}$ when using Euclidean distance, $1$ when using cosine distance. Or, the maximum distance between embeddings should be the distance between opposite unit vectors, $2$ when using Euclidean distance or cosine distance. In some preliminary experiments we trained both orthogonal and opposite, but there did not seem to be much difference, so we went ahead using only orthogonal.

\section{Evaluation Details}
\label{sec:eval_details}
When working on the large datasets, AG News and Yahoo Answers, we used the nmslib library \cite{boytsov2013engineering} and the Hierarchical Navigable World graph (HNSW) method \cite{Malkov2016hierarchical} for approximate nearest neighbours, which builds a slightly different index each time it is run. 

When using f-norm at test time, for each dimension in the embeddings, we draw samples from a standard normal distribution for all training and test embeddings and replace values in the embeddings with the samples which preserve the ranking within the dimension. This procedure will also work if there is only one test point we want to classify.

 \section{Best error rates}
 \label{sec:best_error_rate_details}
We consider which models with which hubness reduction methods give us the lowest error rate when taking the mean over the twelve random seeds. \par 
Models using the largest base model (microsoft-mpnet-base) have the lowest mean error rate. If we take the three best error rates from the three datasets, they are all from models using the largest base model. On 20 Newsgroups using f-norm + MP gives the lowest mean error rate. On AG News, using Euclidean distance during training seemed the best predictor for a low error rate. On Yahoo Answers, using the largest base model, Euclidean distance and z-score normalisation during training gave the four lowest error rates with f-norm + MP being the lowest one. More details below. \par 
On 20 Newsgroups, we find that the ten best mean error rates all used f-norm + MP. This is still true if we only look at models using the medium base model (distilroberta-base) or the small base model (microsoft-MiniLM-L12-H384-uncased). So on this dataset, using the f-norm + MP gives the best mean error rate. \par 
On AG News, the ten best mean error rates all used Euclidean distance for training. The best three also used z-score normalisation. The best three and the seventh best use the largest base model, the rest all use the medium base model. If we only look at models using the medium base model, the best ten still all used Euclidean distance for training, however the best four are no longer z-score normalisation. So on this dataset, using Euclidean distance for training the model seems to be the best predictor for a low mean error rate. \par 
On the Yahoo Answers dataset, the best four all used the largest base model, Euclidean distance and z-score normalisation and with the reductions in order: f-norm + MP, MP, f-norm, none. The numbers four to ten all used the f-norm + MP. If we restrict to the medium base model, the best seven mean error rates are all from models which used Euclidean distance. The best three are after using f-norm + MP, four to six used f-norm, seven used MP and eight to ten used f-norm + MP. So on this dataset, f-norm + MP still gives the best mean error rate, but using the largest base model, Euclidean distance and z-score normalisation gave better error rate than using anything else and then f-norm + MP. 

\section{Differences Between Normalisations During Training}
\label{sec:normalisation_during_training_differences}
Models which use some kind of embedding length normalisation (n; c,n; z) tend to have slightly lower hubness measured with robinhood score than models which do not have this kind of normalisation (none; c). However, there is also a difference between the base models used. On 20 Newsgroups, if we take any two models using microsoft-mpnet-base as base model where one uses some kind of length normalisation (n; c,n; z) and the other does not (none; c), then in about 97.2\% of cases the model not using length normalisation will have the higher robinhood score. If we do the same for two models using microsoft-MiniLM-L12-H384-uncased as base model, the same will be true in about 70.5\% cases. It is still a majority and it is still safe to assume it is a different distribution (Mann-Whitney U test gives a p-value of $3.3 \cdot 10^{-6}$), but it is not as overwhelming. \par 

On 20 Newsgroups for the models using the small and medium base models, the base error rate is worse when using Euclidean distance without than it is with normalisation to unit length. However, for the models using the largest base model (except for one which seems to have failed training), error rates are slightly better for the models without normalisation than with. On AG News and Yahoo Answers, models (medium and large) using Euclidean distance and no normalisation have lower error rates than models with normalisation to unit length and also lower than models using cosine or cosine distance both with and without unit length normalisation. See Appendix \ref{sec:error_none_vs_n} for figures.

\section{Mutual Proximity Implementation}
\label{sec:mp_implementation}
\citep{schnitzer2012local} showed that instead of calculating distances from $x, y$ to all other points in the dataset, it is enough to take a sample of other points, and, if further assuming that distances are approximately normally distributed, stable results can be obtained with as few as 30 samples. \\
We used an implementation of mutual proximity which assumes a normal distribution of distances, since it is faster to compute than the one which allows arbitrary distribution of distances. This is reasonable, since \citep{schnitzer2012local} showed that mutual proximity assuming a normal distribution of distances, had a similar performance to the type without the assumption.

\section{More Skewness Gives Larger Effect of Reduction}
\label{sec:more_skew_more_effect}
On AG News, for the $65$ cases where error rate does not improve when using f-norm + MP, the maximum skewness is $2.58$ and the mean skewness is $2.27$, compared to $9.80$ and $2.80$ for models where error rate improved and $9.80$ and $3.41$ for models where error rate was reduced with at least half a percentage point. There are $96$ models which have a base skewness score of more than $3$ and of these, $91$ cases have an error rate reduction of more than half a percentage point. So the method is more likely to have a larger effect if the base skewness is higher.

\section{Note on SVM}
\label{sec:svm_note}
Using SVM on our trained embeddings for classification on 20 Newsgroups, we saw that in 539 or 506 cases out of 540 when using an rbf or linear kernel, error rate improved from the base case when using f-norm on the embeddings before SVM. Error rate also went down in 455 or 540 cases out of 540 when using an rbf or linear kernel when using f-norm compared to Z-score normalisation before SVM. Exploring this further would be a good direction for future research.

\section{Overview Tables}
\label{sec:overviews}
\begin{figure*}[ht]
\includegraphics[height=0.90\textheight]{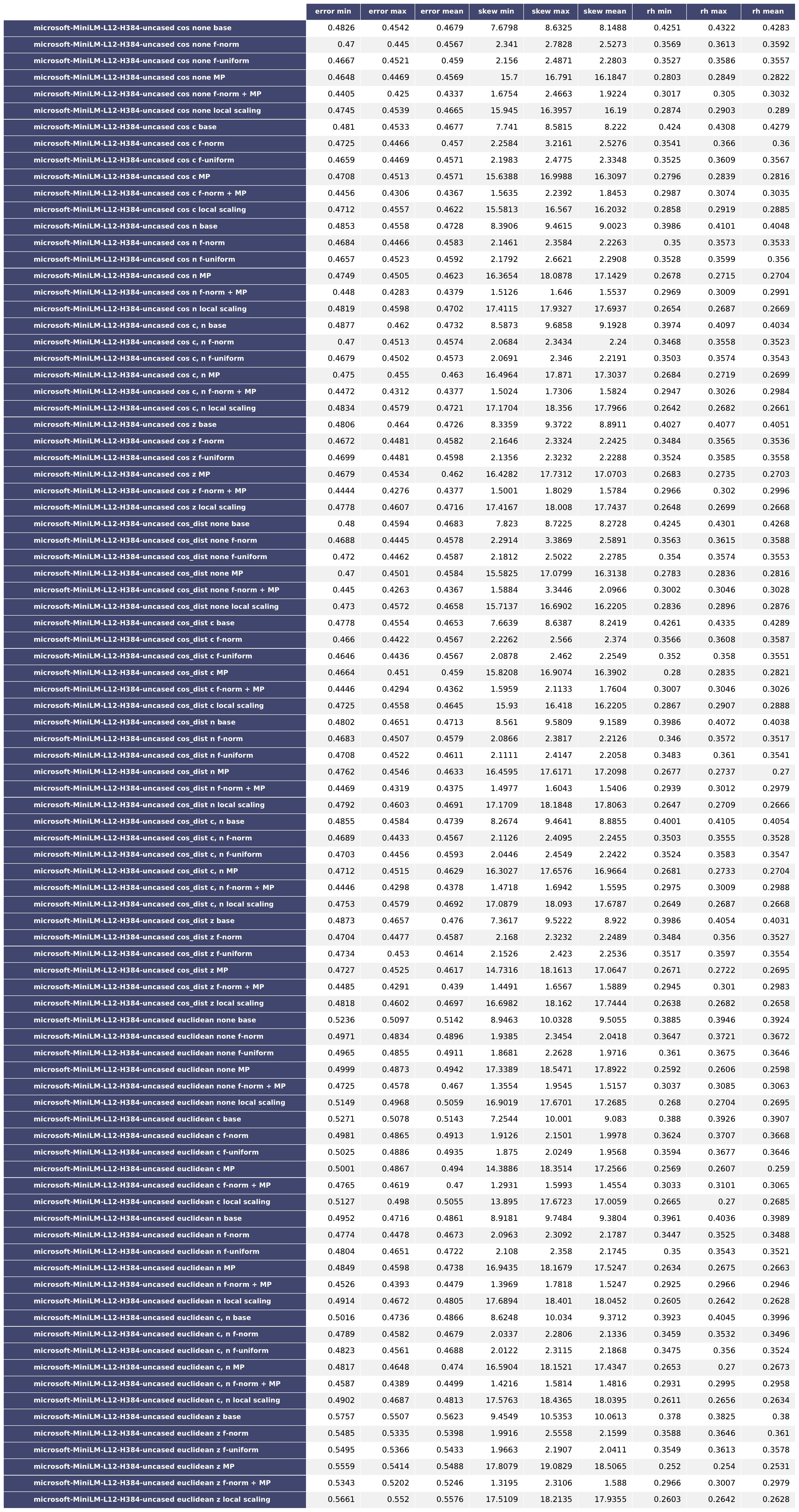}
\caption{\label{fig:newsgroups_results_table_s}Error rate and hubness on the test split of the 20 Newsgroups dataset for the smallest base model tested. Means are over the 12 random seeds. error: Error rate, skew: K-skewness, rh: Robinhood score.}
\end{figure*}

\begin{figure*}[ht]
\includegraphics[height=0.90\textheight]{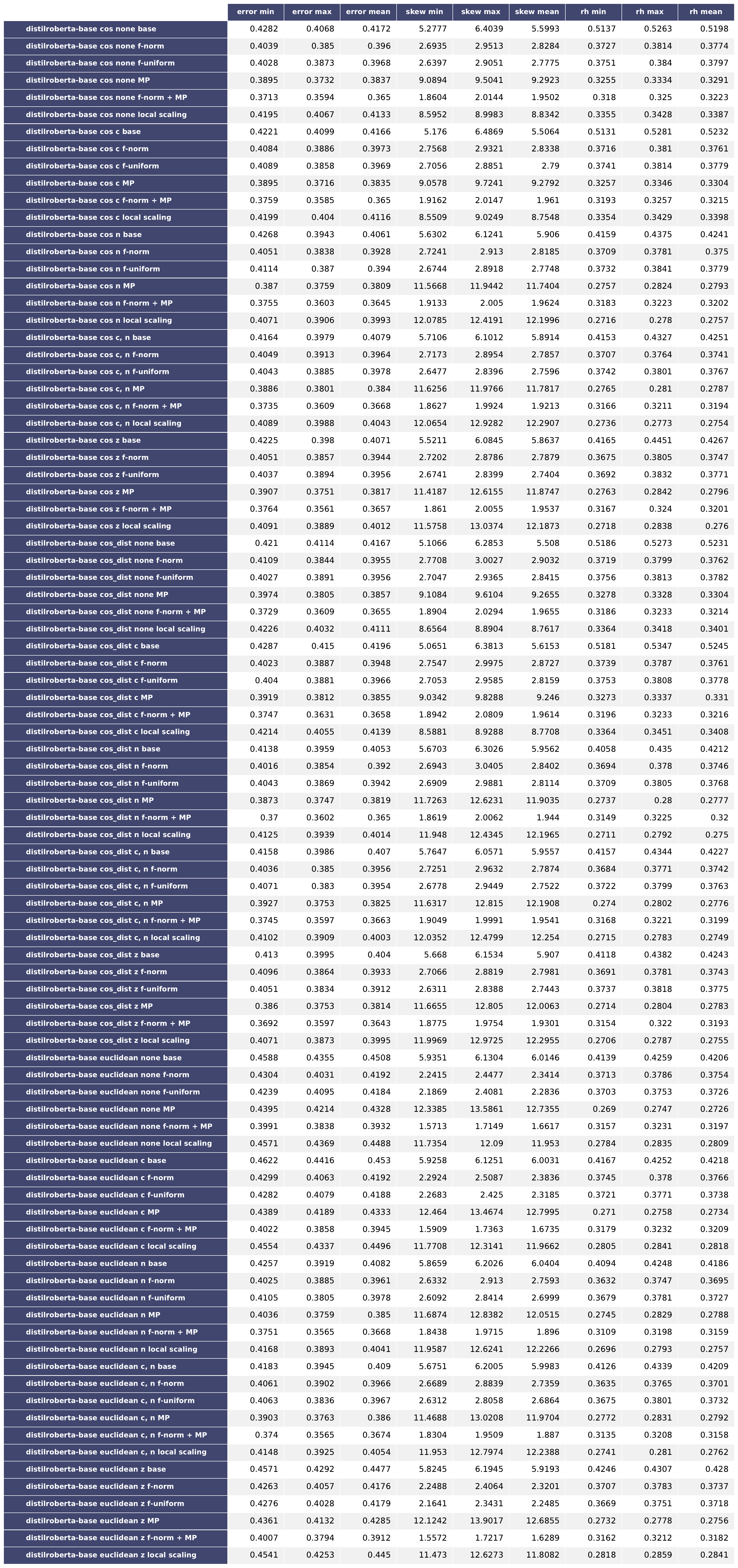}
\caption{\label{fig:newsgroups_results_table_m}Error rate and hubness on the test split of the 20 Newsgroups dataset for the medium base model. Means are over the 12 random seeds. error: Error rate, skew: K-skewness, rh: Robinhood score.}
\end{figure*}

\begin{figure*}[ht]
\includegraphics[height=0.90\textheight]{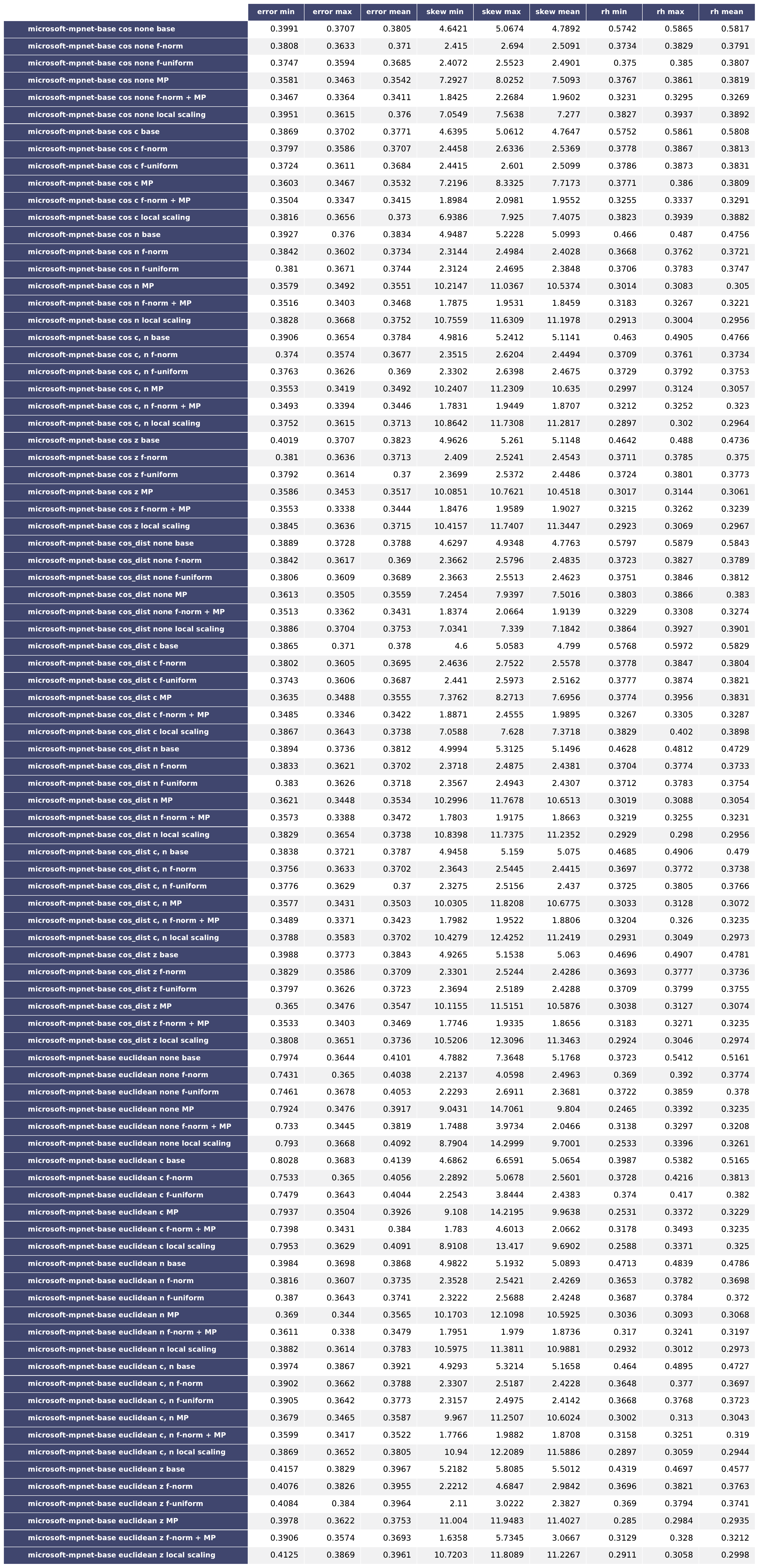}
\caption{\label{fig:newsgroups_results_table_l}Error rate and hubness on the test split of the 20 Newsgroups dataset for the largest base model tested. Means are over the 12 random seeds. error: Error rate, skew: K-skewness, rh: Robinhood score.}
\end{figure*}

\begin{figure*}[ht]
\includegraphics[height=0.90\textheight]{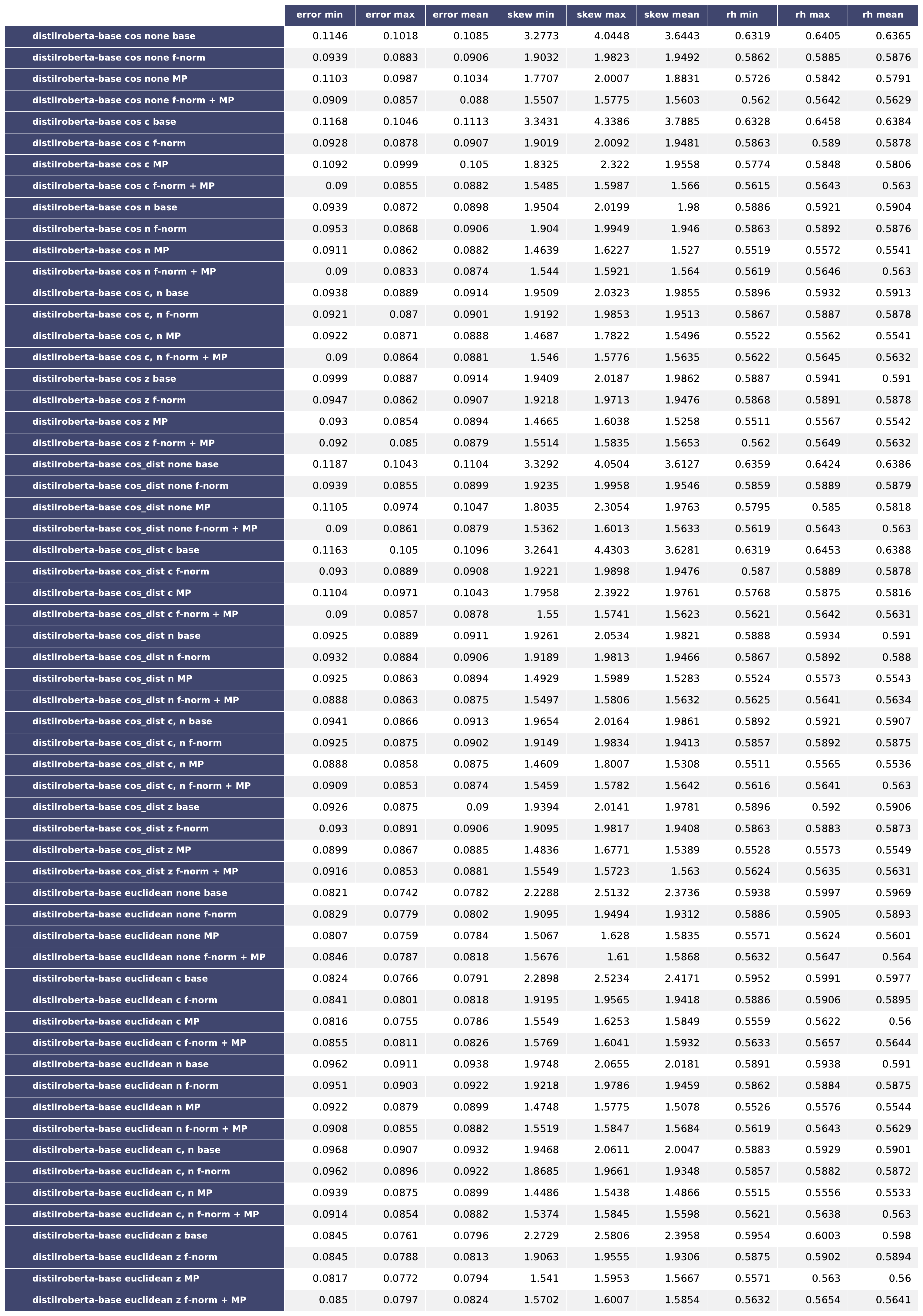}
\caption{\label{fig:ag_news_results_table_m}Error rate and hubness on the test split of the AG News dataset for the medium base model tested. Means are over the 12 random seeds. error: Error rate, skew: K-skewness, rh: Robinhood score.}
\end{figure*}

\begin{figure*}
\includegraphics[height=0.90\textheight]{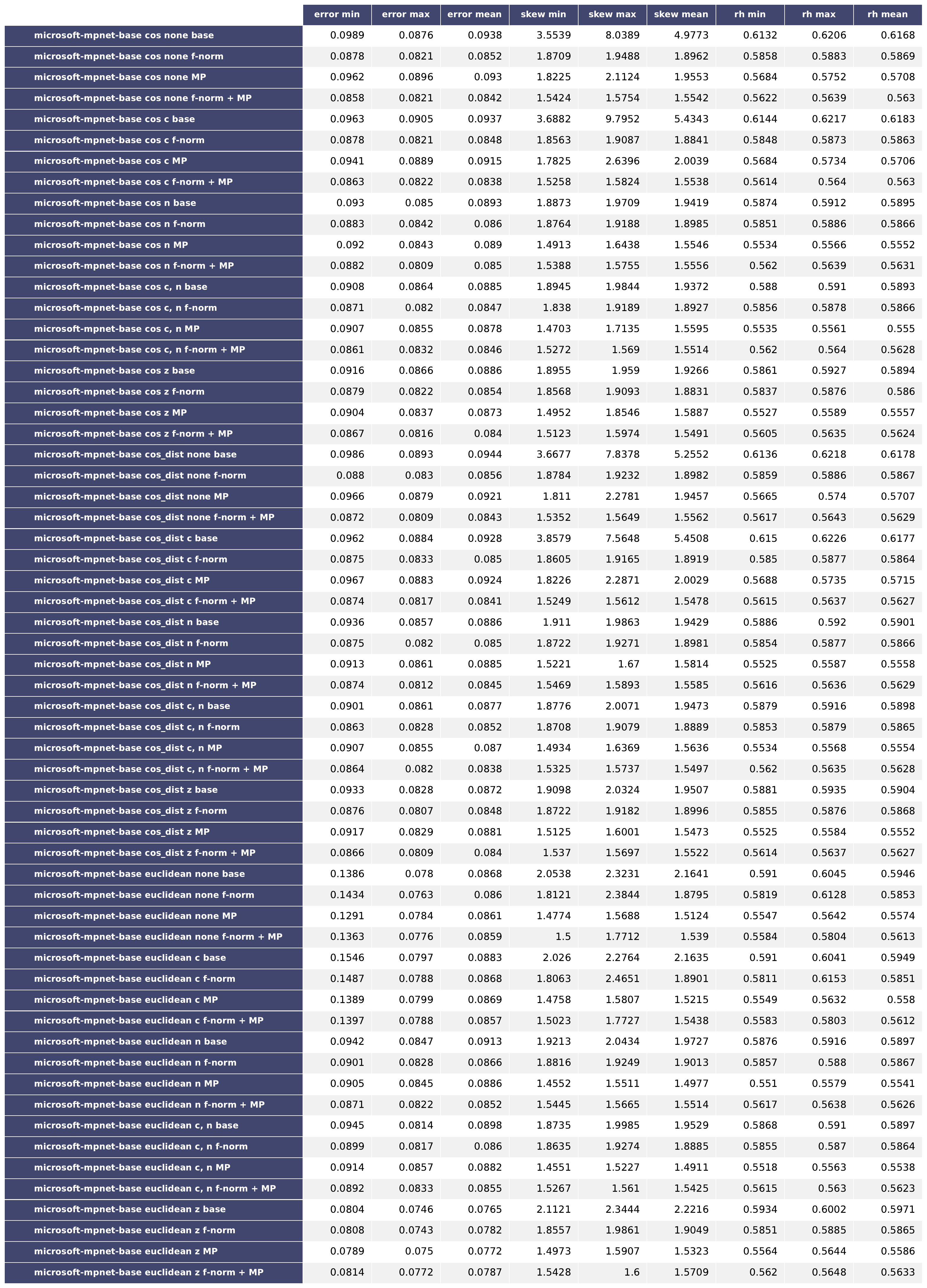}
\caption{\label{fig:ag_news_results_table_l}Error rate and hubness on the test split of the AG News dataset for the largest base model tested. Means are over the 12 random seeds. error: Error rate, skew: K-skewness, rh: Robinhood score.}
\end{figure*}

\begin{figure*}[ht]
\includegraphics[height=0.90\textheight]{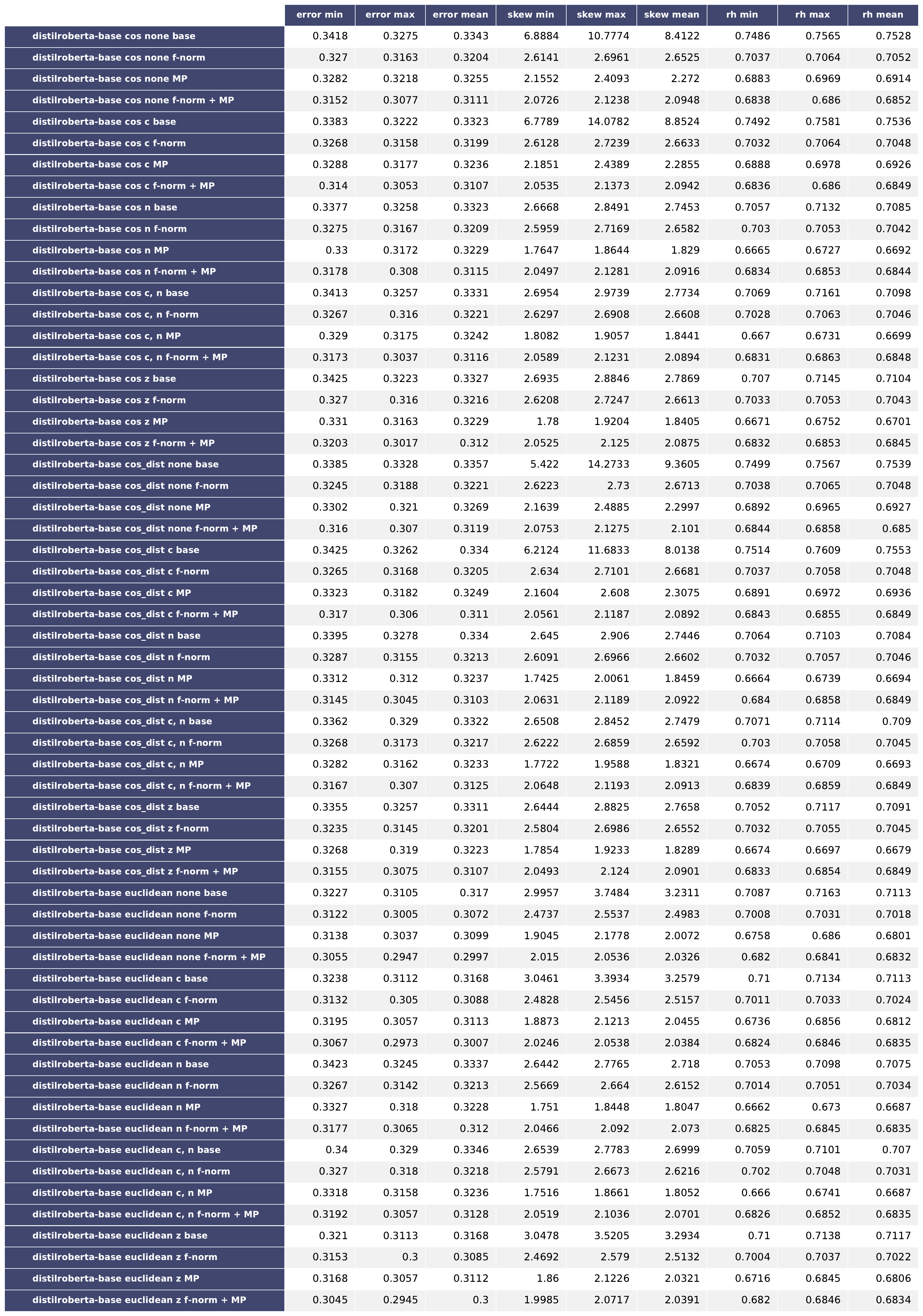}
\caption{\label{fig:yahoo_answers_results_table_m}Error rate and hubness on the test split of the ten percent of the Yahoo Answers dataset for the medium base model tested. Means are over the 12 random seeds. error: Error rate, skew: K-skewness, rh: Robinhood score.}
\end{figure*}

\begin{figure*}[ht]
\includegraphics[height=0.90\textheight]{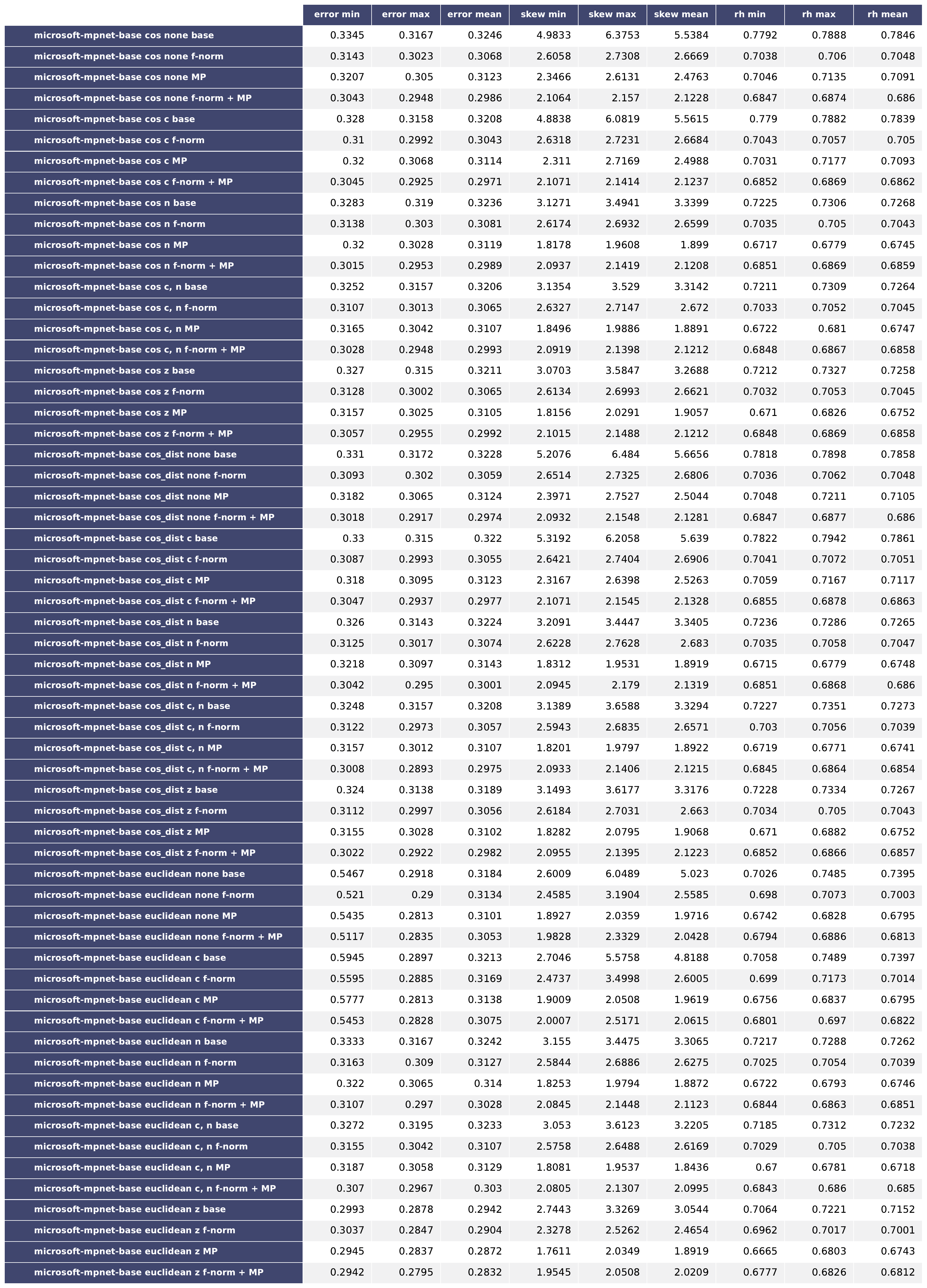}
\caption{\label{fig:yahoo_answers_results_table_l}Error rate and hubness on the test split of the ten percent of the Yahoo Answers dataset for the largest base model tested. Means are over the 12 random seeds. error: Error rate, skew: K-skewness, rh: Robinhood score.}
\end{figure*}

Figures \ref{fig:newsgroups_results_table_s}, \ref{fig:newsgroups_results_table_m}, \ref{fig:newsgroups_results_table_l}, \ref{fig:ag_news_results_table_m}, \ref{fig:ag_news_results_table_l}, \ref{fig:yahoo_answers_results_table_m}, \ref{fig:yahoo_answers_results_table_l} show tables with min, max and mean of error rate(error), k-skewness(skew) and robinhood score(rh) for the 12 random seeds of all our trained model types on the test splits of our datasets.

\end{document}